\documentclass[runningheads]{llncs}

\usepackage{eccv}

\usepackage{eccvabbrv}

\usepackage{graphicx}
\usepackage{booktabs}
\usepackage{multirow}
\usepackage{graphicx}
\usepackage{adjustbox}
\usepackage{listings}
\usepackage{enumitem}
\usepackage{wrapfig}
\usepackage{array}

\usepackage[accsupp]{axessibility}  

\usepackage{hyperref}

\usepackage{orcidlink}

\begin{document}

\title{Freqformer: Image-Demoiréing Transformer via Effective Frequency Decomposition} 

\titlerunning{Freqformer for Image Demoiréing}


\author{Xiaoyang Liu\thanks{Equal contribution} \and Bolin Qiu$^*$ \and Zheng Chen \and Libo Zhu  \and \\ Zihan Zhou  \and Kai Liu \and Jiezhang Cao \and Yulun Zhang\thanks{Corresponding author: Yulun Zhang, yulun100@gmail.com} }

\authorrunning{X. Liu et al.}


\institute{
Shanghai Jiao Tong University
}

\maketitle

\begin{abstract}
  Image demoiréing remains a challenging task due to the complex interplay between texture corruption and color distortions caused by moiré patterns. Existing methods, especially those relying on direct image-to-image restoration, often fail to disentangle these intertwined artifacts effectively. While frequency-aware approaches offer a promising direction, their potential is hindered by the discrete transform (e.g., Haar wavelet or block-based DCT), which may suffer from spatial discontinuity, channel redundancy, and further cause error accumulation during their fixed inverse processes. In this paper, we present Freqformer, a Transformer-based framework specifically designed for image demoiréing through targeted frequency separation. Our method performs an effective frequency decomposition that splits moiré patterns into high-frequency spatially-localized textures and low-frequency scale-robust color distortions, which are then handled by a dual-branch architecture and an asymmetric training scheme tailored to their distinct characteristics. We further propose a learnable Frequency Composition Transform (FCT) module to adaptively fuse the frequency-specific outputs, enabling consistent and high-fidelity reconstruction. To better aggregate the spatial dependencies and the inter-channel complementary information, we introduce a Spatial-Aware Channel Attention (SA-CA) module that refines moiré-sensitive regions without incurring high computational cost. Extensive experiments on various demoiréing benchmarks demonstrate that Freqformer achieves state-of-the-art performance with a compact model size. The code will be made publicly available at~\url{https://github.com/xyLiu339/Freqformer}.
\end{abstract}
\section{Introduction}
\label{sec:intro}

Capturing digital screens with a camera often introduces moiré patterns—irregular, colorful artifacts caused by aliasing between the screen’s subpixel layout and the camera’s color filter array (CFA)~\cite{reason}. These patterns significantly degrade image quality and are difficult to remove due to their complexity and variability. Consequently, single-image demoiréing remains a challenging low-level vision task that continues to attract attention from academia and industry.

Recent advances in deep learning have led to progress in image demoiréing, with CNN-based models achieving promising results~\cite{he2020fhde, he2019mop, zheng2020image,sun2018moire, yuan2019aim, yu2022towards, xu2024image, xiao2024p}. However, most existing approaches treat moiré removal as a holistic restoration task, directly learning to map the input image to a clean output. This straightforward strategy is inherently limited due to the complex nature of moiré patterns: the \textbf{color distortions} caused by moiré patterns are mixed with the original image colors, and the \textbf{corrupted textures} often resemble natural structures. To address this, some recent works have turned to conventional frequency decomposition tools, such as the Haar wavelet~\cite{luowavelet2020cvprw, liu2020wavelet,yeh2024multibranch} or block-based DCT~\cite{zheng2020image, he2020fhde}, to disentangle image components. While frequency-aware methods have shown improved performance, their potential has not been fully explored. For instance, some methods~\cite{luowavelet2020cvprw, liu2020wavelet} simply concatenate wavelet subbands for joint learning, failing to exploit the distinct semantics of different frequency components. Other methods like~\cite{zheng2020image} may fail to capture and eliminate large-scale moiré patterns effectively due to the restricted receptive field.

\begin{wrapfigure}{r}{0.58\linewidth}
\centering
\vspace{-7mm}
\includegraphics[width=\linewidth]{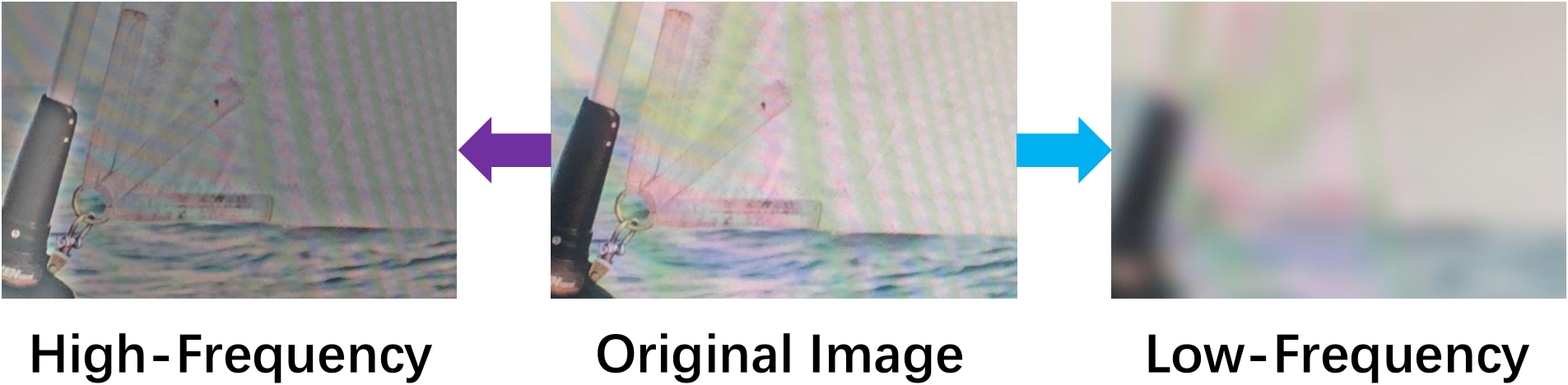}
\vspace{-5mm}
\caption{Our Frequency Decomposition.}
\label{fig:small}
\vspace{-5mm}
\end{wrapfigure}

To overcome these limitations, we introduce a novel frequency decomposition to effectively \textbf{separate moiré textures and color distortions into high-frequency and low-frequency components}, respectively (Fig.~\ref{fig:small}). This enables the design of a dual-branch architecture that learns to handle each frequency band independently, providing more specialized restoration. In addition, benefiting from the favorable properties of our frequency decomposition, the high-frequency moiré textures exhibit strong \textbf{locality}, while the low-frequency components demonstrate \textbf{robustness to scale variations}. Leveraging these characteristics, we adopt an asymmetric training scheme: \textbf{crop-}based training for high-frequency features and \textbf{resize-}based training for low-frequency features. In particular, the resize-based training for the low-frequency component largely avoids detail loss~\cite{he2020fhde} while concentrating on color distortions, making them easier to correct, thereby improving overall performance. Moreover, while traditional frequency-based methods rely on predefined, non-trainable inverse transforms, our decomposition is designed to support a learnable \textbf{F}requency \textbf{C}omposition \textbf{T}ransform (\textbf{FCT}) module, which adaptively fuses the two branches and ensures better consistency in reconstruction.

Additionally, most previous demoiréing methods are built on heavy U-Net-style backbones~\cite{he2019mop,zheng2020image,he2020fhde, yu2022towards} (10M$\sim$50M), which are effective but computationally inefficient, especially for ultra-high-resolution images (\textit{e.g.}, 4K). Recent findings~\cite{xiao2024p} have shown that moiré artifacts exhibit variance across color channels, which motivates more refined attention mechanisms. To this end, we introduce a \textbf{S}patial-\textbf{A}ware \textbf{C}hannel \textbf{A}ttention (\textbf{SA-CA}) module that enhances spatial and channel-wise feature modeling while maintaining a compact design. Unlike standard attention modules, SA-CA selectively emphasizes moiré-prone regions and channels without incurring excessive computational overhead, making our model suitable for high-resolution inference and deployment on edge devices.

By integrating the above components, we propose \textbf{Freqformer}, a hierarchical dual-branch Transformer tailored for image demoiréing. Freqformer leverages frequency feature separation, adaptive fusion, and efficient attention mechanisms to remove moiré artifacts. It achieves state-of-the-art performance on various high-resolution image demoiréing benchmarks, while maintaining a compact model size. Our main contributions are summarized as follows:
\vspace{-3mm}
\begin{itemize}[left=1em, itemsep=-1pt]
    \item We introduce a novel frequency decomposition strategy to effectively separate moiré patterns into high-frequency (texture) and low-frequency (color) components, enabling targeted learning for each.
    \item We design a dual-branch, asymmetric architecture that separately processes high- and low-frequency components using crop- and resize-based training strategies, enabling specialized feature extraction tailored to different aspects of moiré patterns. We also introduce a learnable FCT module that adaptively fuses the outputs from both branches.
    \item We propose Freqformer, with a lightweight SA-CA module to enhance both spatial and channel-wise feature interactions. Extensive experiments on image demoiréing benchmarks show that Freqformer achieves state-of-the-art performance with a small number of parameters.
\end{itemize}

\vspace{-5mm}
\section{Related Works}
\label{sec:formatting}
\vspace{-2mm}
\subsection{Demoir\'{e}ing Methods}
\vspace{-1mm}
\paragraph{Image demoiréing.}Early image demoiréing methods focused on specific moiré patterns. These approaches, such as smooth filtering~\cite{siddiqui2009hardware}, spectral analysis~\cite{sidorov2002suppression}, and image decomposition~\cite{liu2015moire}, rely heavily on handcrafted features and mathematical models and often fail to generalize effectively to diverse moiré patterns and varying image conditions. In recent years, deep learning-based methods~\cite{he2019mop, zheng2020image,sun2018moire, yuan2019aim, zhang2023real, peng2024image} have emerged as a powerful tool for image demoiréing. Yu et al.~\cite{yu2022towards} proposed ESDNet, the first lightweight model for ultra-high-resolution image demoiréing.  MoiréDet~\cite{yang2023doing} further extracts moiré edge maps from images with moiré patterns using a triplet generation strategy. Xiao et al.~\cite{xiao2024p} developed a patch bilateral compensation network, leveraging the green channel's resistance to moiré patterns. Peng et al.~\cite{peng2024image} present DMSFN, a network using dilated-dense attention and multiscale feature interaction for effective moiré pattern removal, along with a moiré data augmentation technique.

\vspace{-3mm}
\paragraph{Video demoiréing.} Video demoiréing poses added challenges due to the need for temporal coherence. Dai et al.~\cite{dai2022video} introduced the first hand-held video demoiréing dataset and a baseline model using relation-based temporal consistency loss. Cheng et al.~\cite{cheng2023recaptured} extended image and video demoiréing to Raw inputs with color-separated and spatial modulations. Xu et al.~\cite{xu2024direction} proposed DTNet, a unified framework that performs moiré removal, alignment, color correction, and detail refinement via directional DCT modes and temporal-guided bilateral learning.

\vspace{-3mm}
\paragraph{Raw domain demoiréing.}The Raw domain offers high-bit-depth data with reduced ISP interference, making it well-suited for low-level tasks. This is especially true for demoiréing, where moiré patterns are simpler in Raw~\cite{yue2022recaptured}, free from ISP-induced nonlinearities. RDNet~\cite{yue2022recaptured} first tackles Raw image demoiréing with an encoder-decoder and class-specific learning. RawVDemoiré~\cite{cheng2023recaptured} addresses video demoiréing via temporal alignment. Recent work~\cite{xu2024image} jointly leverages Raw and sRGB inputs to learn device-dependent ISP models for restoration. Despite these advantages, Raw-domain demoiréing is inherently sensor-dependent and struggles with the subsequent Raw-to-sRGB conversion, where inaccurate ISP approximations can lead to color shifts. Thus, we perform demoiréing directly in the RGB domain to better approximate the mapping to real-world scenes.

\vspace{-3mm}
\paragraph{Moir\'{e} removal dataset.}Sun et al.~\cite{sun2018moire} introduced a large-scale benchmark dataset (TIP-2018) containing over 100,000 image pairs. Zheng et al.~\cite{zheng2020image} proposed the AIM2019 image demoiréing challenge dataset (LCDMoiré) in 2020. LCDMoiré contains only text scenes, while TIP-2018 does not include text scenes but has a limited resolution of 256$\times$256. FHDMi dataset, presented by He et al.~\cite{he2020fhde}, provides high-resolution images with more diverse and complex moiré patterns. More recently, Yu et al.~\cite{yu2022towards} took a further step by constructing the first large-scale real-world Ultra-High-Definition demoiréing dataset (UHDM), containing real-world 4K resolution image pairs with more challenging and practical scenarios. 
Yue et al.~\cite{yue2022recaptured} built the first well-aligned raw moiré image dataset (TMM-2022) by pixel-wise alignment between the recaptured images and source ones.

\vspace{-2mm}
\subsection{Frequency-based Methods}
\vspace{-2mm}
Wavelet-based methods, due to their ability to decompose an image into multiple frequency bands, have been widely applied in some computer vision tasks, including classification\cite{fujieda2017wavelet}\cite{li2020wavelet}\cite{oyallon2017scaling}\cite{williams2018wavelet}, network compression\cite{gueguen2018faster}\cite{levinskis2013convolutional}, face aging\cite{liu2019attribute}, super-resolution\cite{huang2017wavelet}\cite{liu2018multi}, style transfer\cite{yoo2019photorealistic}, etc. In recent years, wavelet-based methods have emerged as powerful tools for image demoiréing. Liu et al.~\cite{liu2020wavelet} proposed a wavelet-based dual-branch network (WDNet) that operates in the wavelet domain to effectively separate moiré patterns from image content by leveraging dense and dilated convolution modules. Luo et al.~\cite{luowavelet2020cvprw} propose AWUDN, a deep network that combines wavelet-domain feature mapping with a domain adaptation mechanism. Nevertheless, these two methods simply mix different frequency features without exploiting the hierarchical frequency features of wavelet decomposition. Further, Yeh et al.~\cite{yeh2024multibranch} introduced a multibranch wavelet-based network (MBWDN) that decomposes moiré images into multiple sub-band images using wavelet transform and processes them through different branches. However, due to its multi-level structure, the method incurs high computational cost. Additionally, the Haar transform only partially separates moiré patterns, limiting its effectiveness for high-definition or complex cases.

Besides wavelets, other frequency tools like DCT have also been explored. MBCNN~\cite{zheng2020image} explicitly removes moiré patterns by configuring the Inverse DCT (IDCT) matrix and learnable weights as a bandpass filter. Nevertheless, its effectiveness is bottlenecked by a limited receptive field, hindering the removal of large-scale moiré structures. Meanwhile, FHDe²Net~\cite{he2020fhde} uses DCT on the luminance channel to disentangle high-frequency details from moiré interference. However, utilizing DCT primarily to compensate for downsampling-induced detail loss significantly increases system design complexity.

\begin{figure*}[!ht]
\scriptsize
\centering
\begin{tabular}{c}
\hspace{-0.15cm}
\begin{adjustbox}{valign=t}
\begin{tabular}{cccccc}
\raisebox{2mm}{\includegraphics[width=0.088\textwidth,height=0.1\textwidth]{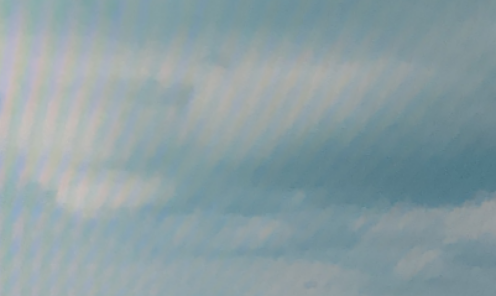}} \hspace{-1.5mm} &
\includegraphics[width=0.15\textwidth,height=0.12\textwidth]{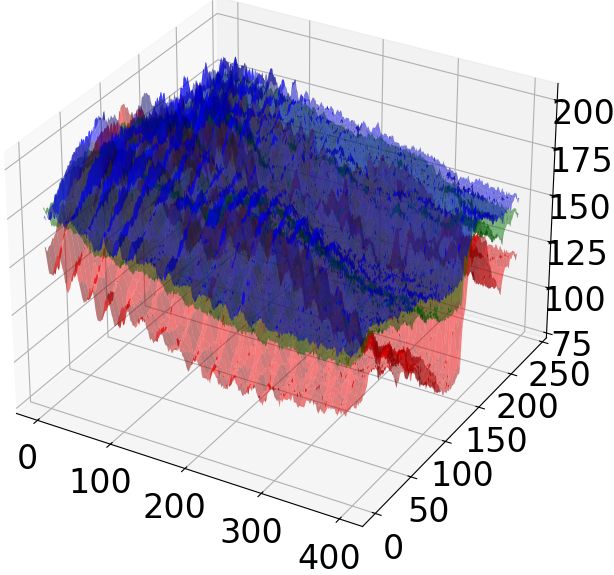} \hspace{-1.5mm} &
\includegraphics[width=0.22\textwidth,height=0.12\textwidth]{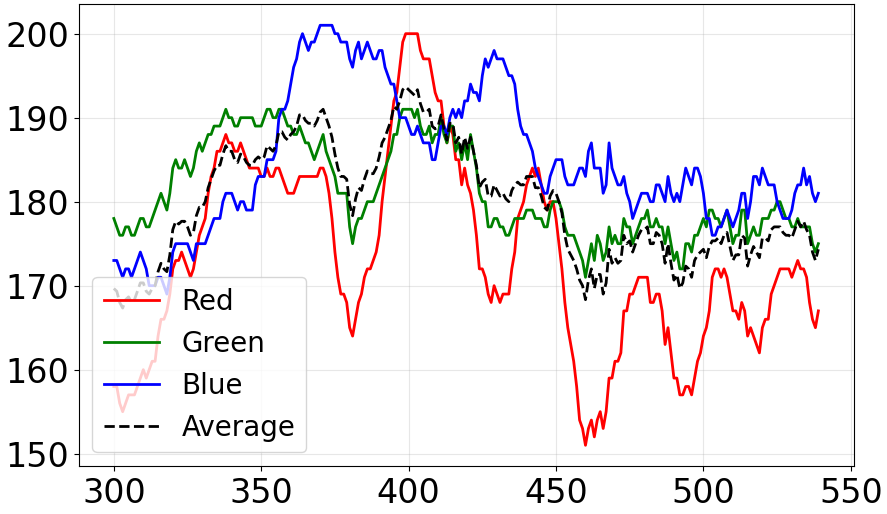} \hspace{-1.5mm} &
\raisebox{2mm}{\includegraphics[width=0.088\textwidth,height=0.1\textwidth]{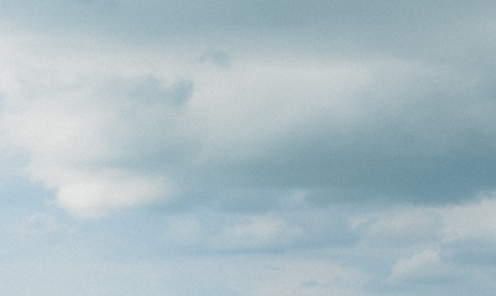}} \hspace{-1.5mm} &
\includegraphics[width=0.15\textwidth,height=0.12\textwidth]{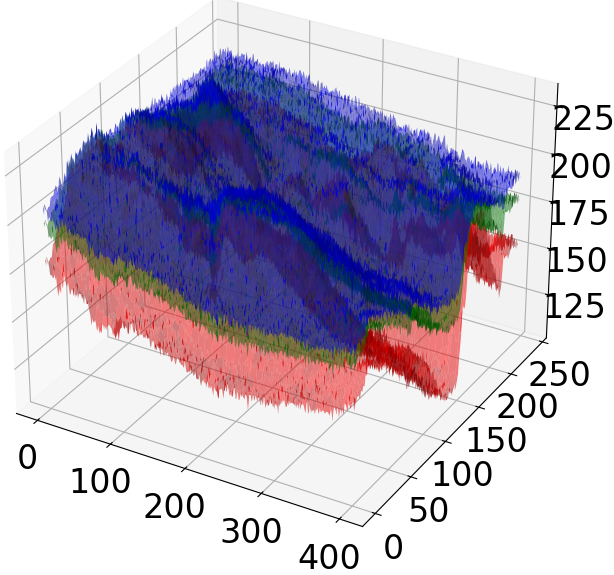} \hspace{-1.5mm} &
\includegraphics[width=0.22\textwidth,height=0.12\textwidth]{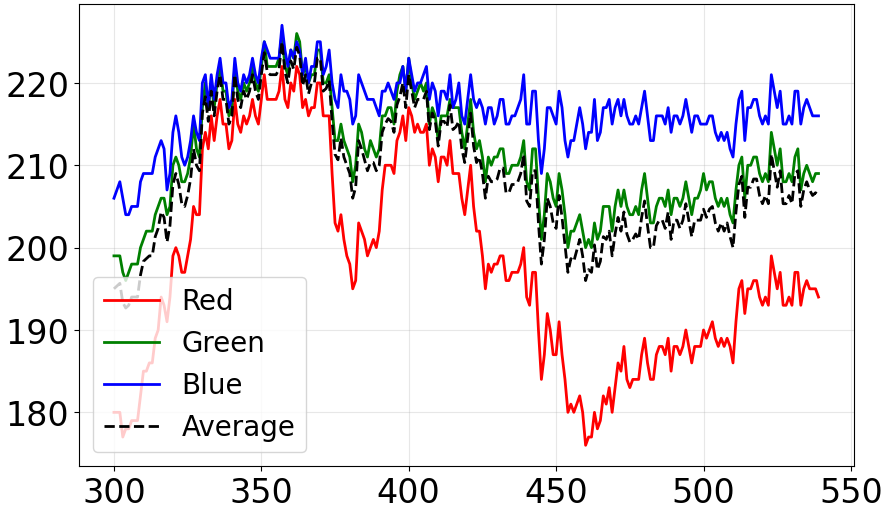} \hspace{-1.5mm} 
\\
Moiré Image \hspace{-1.5mm}
 &
3D RGB \hspace{-1.5mm}  &
2D RGB \hspace{-1.5mm} &
Clean Image \hspace{-1.5mm} &
3D RGB \hspace{-1.5mm}  &
2D RGB \hspace{-1.5mm}
\\
\raisebox{2mm}{\includegraphics[width=0.088\textwidth,height=0.1\textwidth]{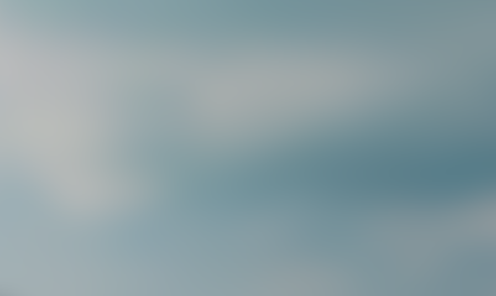}} \hspace{-1.5mm} &
\includegraphics[width=0.15\textwidth,height=0.12\textwidth]{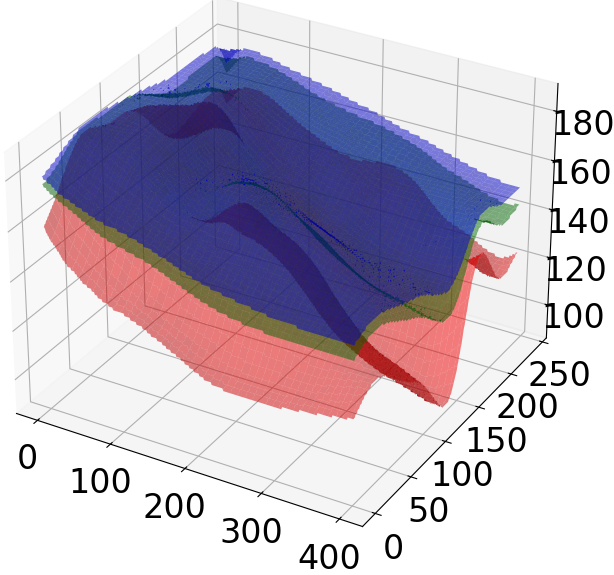} \hspace{-1.5mm} &
\includegraphics[width=0.22\textwidth,height=0.12\textwidth]{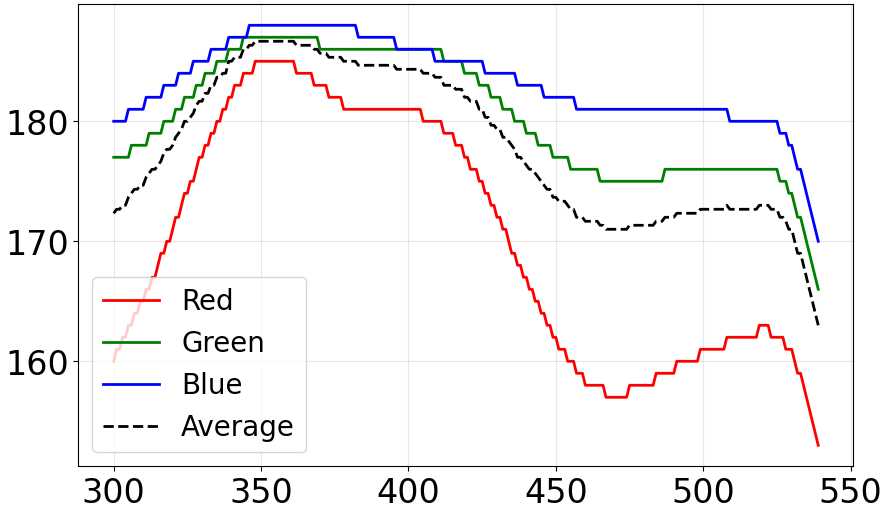} \hspace{-1.5mm}  &
\raisebox{2mm}{\includegraphics[width=0.088\textwidth,height=0.1\textwidth]{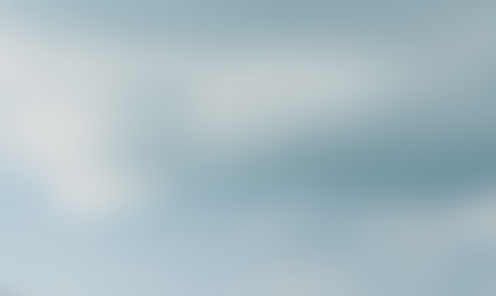}} \hspace{-1.5mm} &
\includegraphics[width=0.15\textwidth,height=0.12\textwidth]{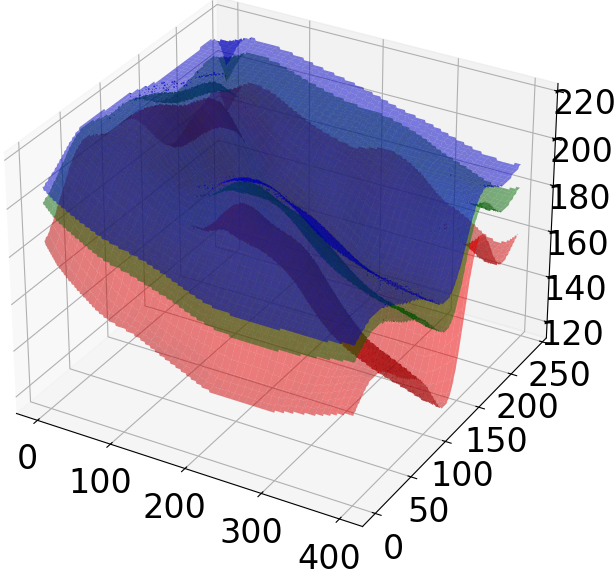} \hspace{-1.5mm} &
\includegraphics[width=0.22\textwidth,height=0.12\textwidth]{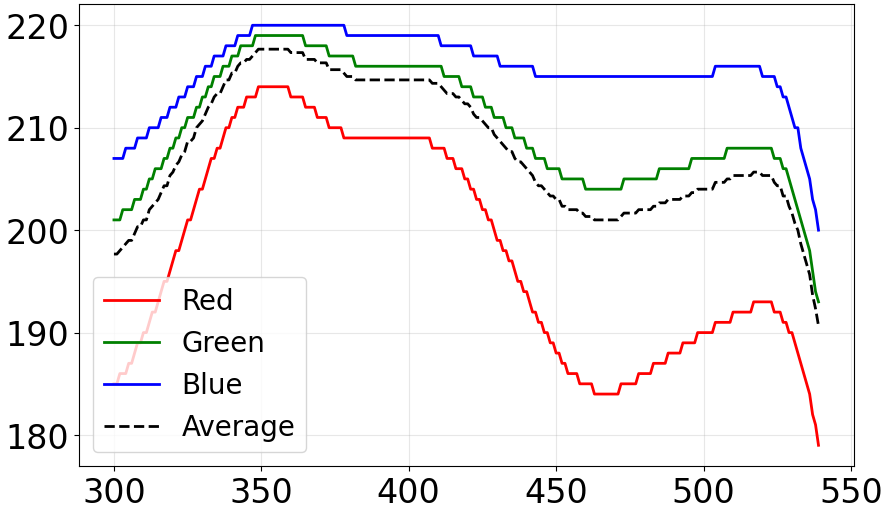} \hspace{-1.5mm}  
\\ 
Moiré Low \hspace{-1.5mm}
 &
3D RGB \hspace{-1.5mm}  &
2D RGB \hspace{-1.5mm} &
Clean Low \hspace{-1.5mm} &
3D RGB \hspace{-1.5mm}  &
2D RGB \hspace{-1.5mm} 
\\
\raisebox{2mm}{\includegraphics[width=0.088\textwidth,height=0.1\textwidth]{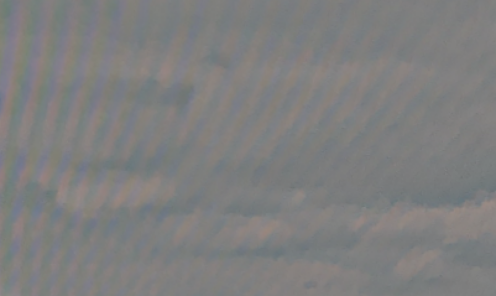}} \hspace{-1.5mm} &
\includegraphics[width=0.15\textwidth,height=0.12\textwidth]{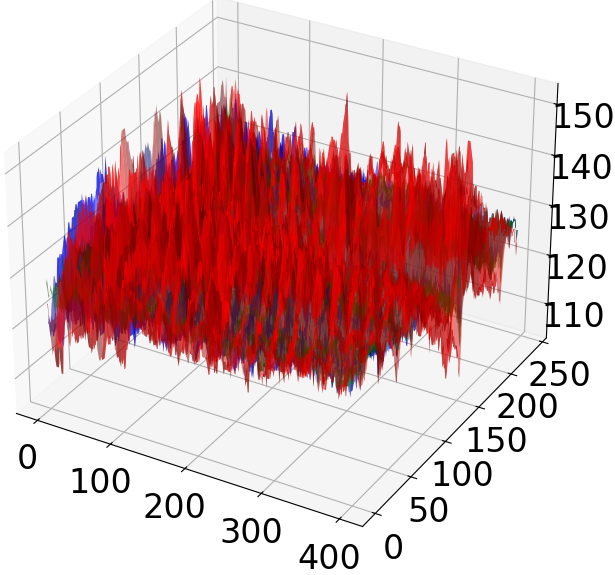} \hspace{-1.5mm} &
\includegraphics[width=0.22\textwidth,height=0.12\textwidth]{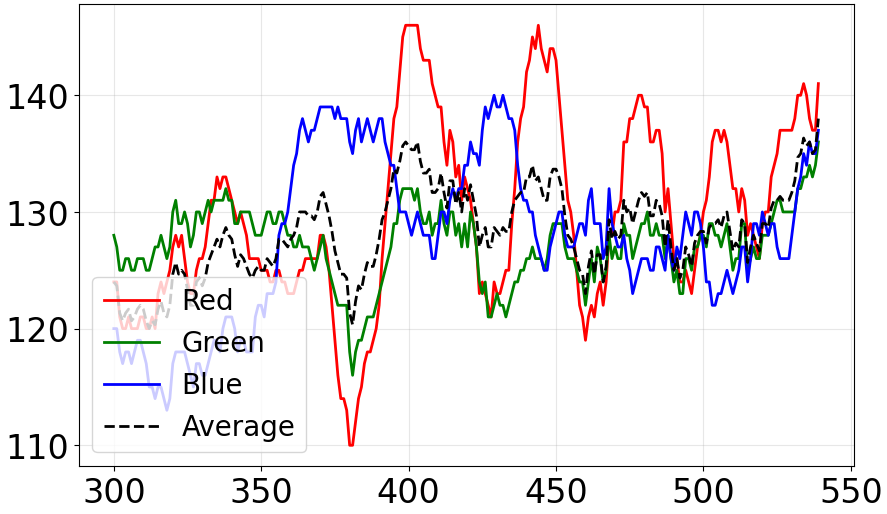} \hspace{-1.5mm} & \raisebox{2mm}{\includegraphics[width=0.088\textwidth,height=0.1\textwidth]{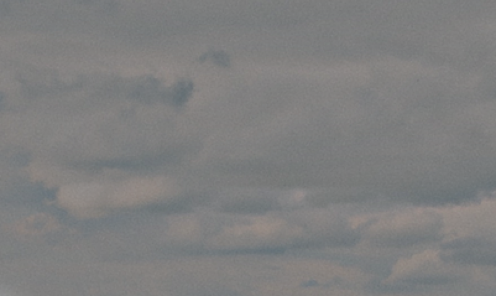}} \hspace{-1.5mm} &
\includegraphics[width=0.15\textwidth,height=0.12\textwidth]{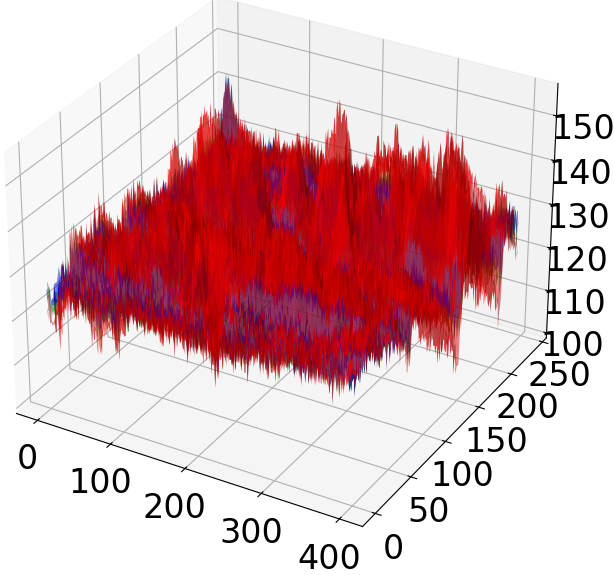} \hspace{-1.5mm} &
\includegraphics[width=0.22\textwidth,height=0.12\textwidth]{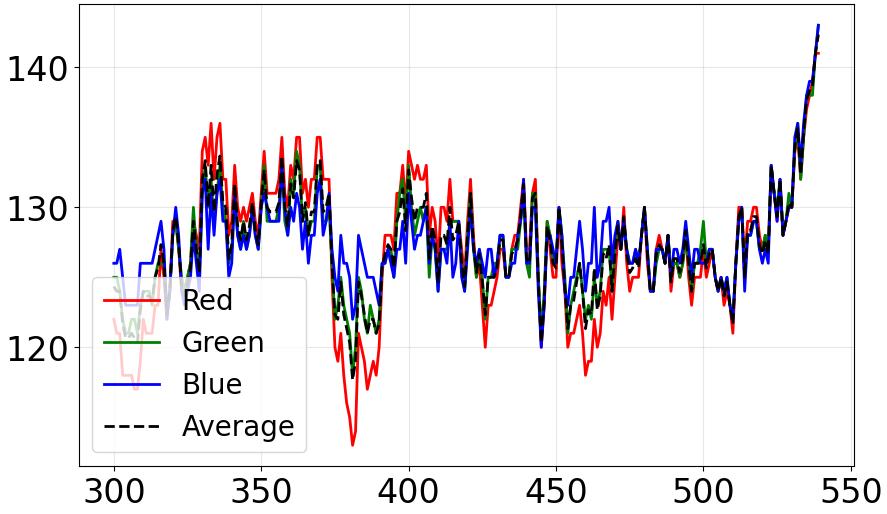} \hspace{-1.5mm}   
\\ 
Moiré High \hspace{-1.5mm}
 &
3D RGB \hspace{-1.5mm}  &
2D RGB \hspace{-1.5mm} &
Clean High \hspace{-1.5mm} &
3D RGB \hspace{-1.5mm}  &
2D RGB \hspace{-1.5mm}
\\
\end{tabular}
\end{adjustbox}

\end{tabular}
\vspace{-1mm}
\caption{Visualization of the frequency-decomposed images (zoom in for better analysis). The left three columns show the moiré images along with their 3D and 2D (a column of pixels) visualizations, while the right three columns display the clean images. The first, second, and third rows represent the original images, low- and high-frequency components, respectively. The low-frequency curves of moiré and clean images exhibit similar trends, primarily reflecting color distortions. The high-frequency curves differ significantly in shape, indicating structural differences caused by moiré patterns.}
\vspace{-3mm}
\label{fig:first_comp}
\end{figure*}


\section{Method}
\vspace{-3mm}
Our whole pipeline is shown in Fig.~\ref{fig:pipe_p1}. Given a moiré image $\boldsymbol{I_m}$, we first apply a frequency decomposition (Fig.~\ref{fig:first_comp}, Sec.~\ref{sec:fc}) to obtain a high-frequency component $\boldsymbol{I_h}$ and a low-frequency component $\boldsymbol{I_l}$, both retaining the original spatial resolution. During training, the high-frequency branch adopts a cropping strategy, while the low-frequency branch uses resizing (Sec.~\ref{sec:sa-ca}). During testing, the high branch directly processes the full-resolution input, while the low branch continues to operate with the resize-based strategy.

The two branches process $\boldsymbol{I_h}/\boldsymbol{I_l}$ using the encoder-decoder architecture. Each encoder and decoder block is a Spatial-Aware Channel Attention (SA-CA) module (Sec.~\ref{sec:sa-ca}). Moreover, a hierarchical feature fusion strategy is adopted in the high-frequency branch, where multi-scale features are concatenated with minimal overhead. 

The network is trained in two phases (Sec.~\ref{sec:loss}). First, the high- and low-frequency branches are trained independently. Then, both branches and the learnable Frequency Composition Transform (FCT) (Sec.~\ref{sec:fc}) are jointly fine-tuned to fuse their features and reconstruct a cleaner output.

\subsection{Frequency Decomposition and Learnable FCT}
\label{sec:fc}

As shown in Fig.~\ref{fig:first_comp}, we decompose both the moiré-contaminated image and the clean image into high- and low-frequency components and visualize the 2D/3D RGB curves. It is evident that the moiré patterns are predominantly preserved in the high-frequency part, while the low-frequency component exhibits minimal moiré textures but suffers from noticeable color distortions. Notably, although the low-frequency curve of the moiré image follows a similar overall trend to that of the clean image, there is a significant deviation in magnitude. This discrepancy can be attributed to two main factors. First, the presence of moiré patterns inherently distorts the original color information during the frequency decomposition, which complicates the accurate recovery of low-frequency components and remains a significant challenge. Second, the dataset construction process itself introduces severe color mismatches — or more precisely, color shifts (also mentioned in~\cite{reason, zheng2020image, xiao2024p}) — due to varying external conditions such as lighting during the moiré image capture in the dataset construction process. Therefore, following the frequency decomposition, we design a dual-branch approach (will be detailed in Sec.~\ref{sec:sa-ca}): a high-frequency branch dedicated to restoring fine textures affected by moiré patterns, and a low-frequency branch focused on recovering accurate color information, addressing both moiré-induced distortions and dataset-related color shifts.

\begin{wrapfigure}{r}{0.58\linewidth}
\centering
\vspace{-9mm}
\includegraphics[width=\linewidth]{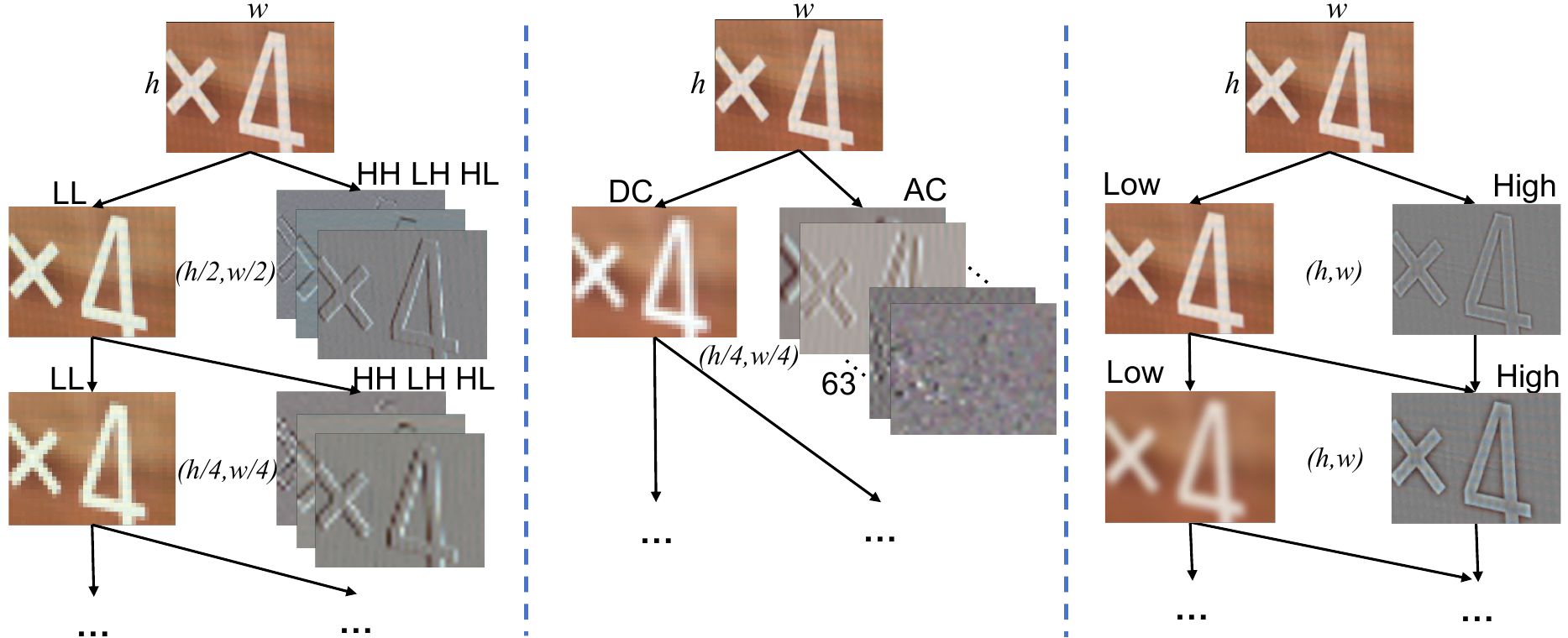}
\vspace{-5mm}
\caption{Frequency decomposition methods. Left: Haar DWT; Middle: Block DCT; Right: Ours.}
\label{fig:haardct}
\vspace{-5mm}
\end{wrapfigure}

While frequency decomposition is crucial for demoiréing, conventional tools like the Discrete Wavelet Transform (DWT, e.g., multi-level Haar)~\cite{luowavelet2020cvprw, liu2020wavelet,yeh2024multibranch} and Discrete Cosine Transform (DCT)~\cite{zheng2020image, he2020fhde} are fundamentally ill-suited for our design needs due to several inherent limitations:
\begin{itemize}[left=0.5em, itemsep=-1pt]
    \item \textbf{Lack of Spatial Smoothness and Error Accumulation:} Both DWT and DCT yield discrete representations lacking spatial smoothness (see the left LL and middle DC figures in Fig.~\ref{fig:haardct}). With rigidly fixed inverse transforms, subtle frequency-domain restoration errors inevitably accumulate during reconstruction, causing severe pixel-level and structural distortions.
    \item \textbf{Resolution Degradation and Shift Variance:} Multi-level recursive decomposition aggressively downsamples spatial resolution (e.g., Haar, from H$\times$W to H/2$^i$$\times$W/2$^i$, and similarly for DCT), risking secondary aliasing. Furthermore, standard downsampled transforms (e.g., Haar) are shift-variant, yielding highly unstable representations for phase-sensitive moiré artifacts.
    \item \textbf{Incomplete Decoupling and Block Artifacts:} DWT's rapid resolution drop often leaves high-frequency moiré entangled within low-frequency components. Similarly, patch-based DCT assumes local signal stationarity; applying it to non-stationary, curved moiré patterns misses global structures and may introduce block artifacts.
    \item \textbf{Architectural Incompatibility and Channel Redundancy:} DWT scatters features across directional sub-bands (HH, HL, LH, LL), while block DCT splits signals into up to 64 isolated channels. Feeding such highly fragmented, redundant representations into neural networks drastically inflates computational overhead without proportional restoration gains.
\end{itemize}
To overcome these issues, we utilize an anti-aliasing and smoothness-preserving frequency transform with a learnable inverse process, which better separates moiré artifacts and supports a more training-friendly Transformer framework.
Inspired by Wang et al.~\cite{ourwavelet},  our decomposition design is as follows. Given a moiré image $\boldsymbol{I_m}$, we utilize convolution with kernel $\boldsymbol{k}$, which is defined as
\begin{equation}
  \boldsymbol{k}=\left[\begin{array}{ccc}
1 / 16 & 1 / 8 & 1 / 16 \\
1 / 8 & 1 / 4 & 1 / 8 \\
1 / 16 & 1 / 8 & 1 / 16
\end{array}\right].
\end{equation}
The recursive decomposition is comprised of $\boldsymbol{L}$-level. For level $\boldsymbol{i}$'s decomposition process ($\boldsymbol{i} = 1,2, ..., \boldsymbol{L}$), the decomposition outcome $\boldsymbol{I^{i}_l}$ and $\boldsymbol{I^{i}_h}$ is calculated as
\begin{align}
    & \boldsymbol{I^{i}_l}=Conv\left(\boldsymbol{I^{i-1}_l}, \boldsymbol{k}, i\right), \\ \label{eq: fd} & \boldsymbol{I}^i_h = \boldsymbol{I}^{i-1}_h +\boldsymbol{I}^{i-1}_l-\boldsymbol{I}^i_l = \boldsymbol{I}^0_l-\boldsymbol{I}^i_l ,
\end{align}
where $\boldsymbol{I}^0_l = \boldsymbol{I_m}$, $\boldsymbol{I}^0_h = \boldsymbol{0}$ and $Conv$ represents the convolution operation with a dilation rate of $2^i$. Ultimately, the $\boldsymbol{L}$-level decomposition results of image $\boldsymbol{I}$ are obtained as $\boldsymbol{I_l}=\boldsymbol{I}^{L}_l$ and $\boldsymbol{I_h}=\boldsymbol{I}^{L}_h$.

\begin{figure*}[t]
    \centering
    \includegraphics[width=0.98\linewidth]{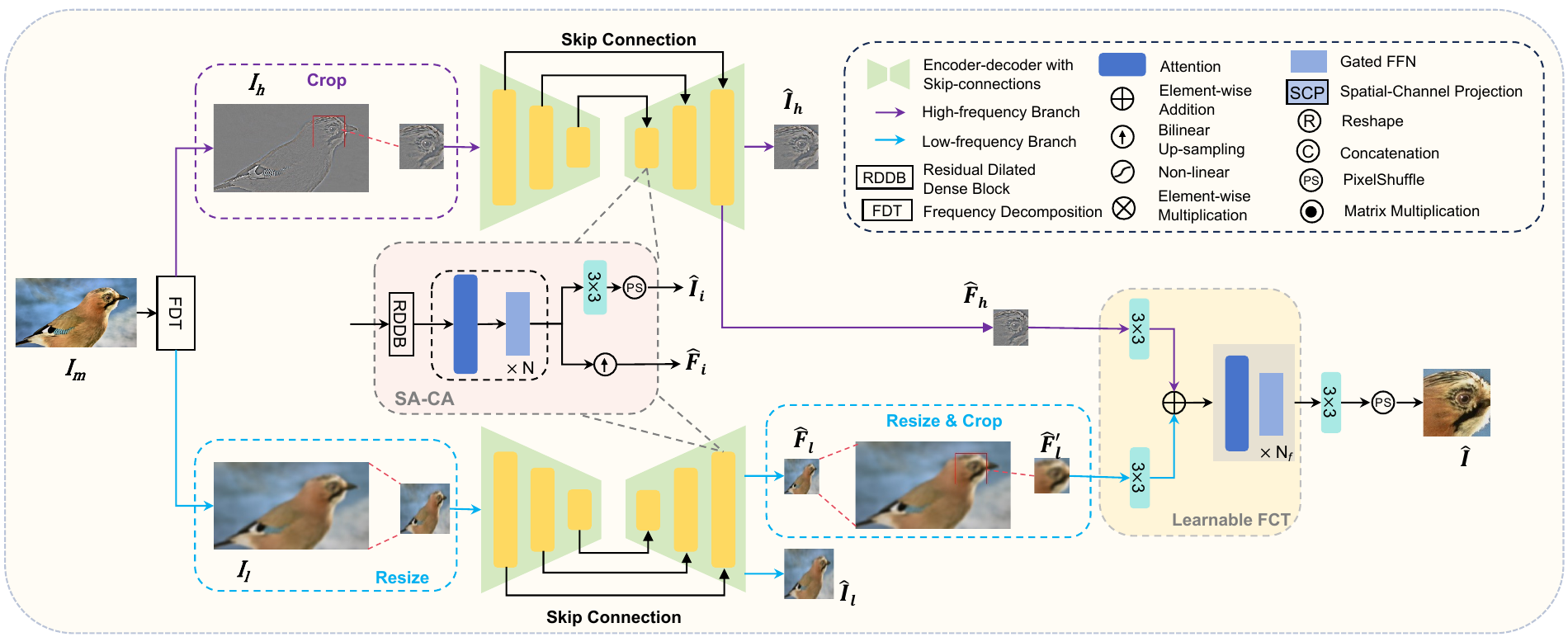}
    \caption{The whole training pipeline. The figure shows the dual-branch learning with crop/resize strategy and a learnable FCT to aggregate the two branches' features.}
    \label{fig:pipe_p1}
    \vspace{-4mm}
\end{figure*}

After the dual-branch demoiré process, we can get the clean representation $\hat{\boldsymbol{I}_l}$ and $\hat{\boldsymbol{I}_h}$. Normally, according to Eq.~\ref{eq: fd}, the final output is calculated as
\begin{equation}
    \boldsymbol{\hat{I}} = \boldsymbol{\hat{I}_l} + \boldsymbol{\hat{I}_h},
\end{equation}
But this traditional composition process cannot fuse the two branches effectively, as the reconstruction error in the same area on the low and high frequency will further lead to more severe distortions. So we instead design a learnable Frequency Composition Transform (FCT) which operates on the feature space. According to Fig.~\ref{fig:pipe_p1}, we extract the feature before the convolution that projects the feature dimension to image dimension and the PixelShuffle, and denote it as $\hat{\boldsymbol{F}_l}$ and $\hat{\boldsymbol{F}_h}$. The learnable FCT module is defined as
\begin{align}
    & \boldsymbol{F} = Conv (\boldsymbol{\hat{F}_l}) + Conv (\boldsymbol{\hat{F}_h}), \\
    & \boldsymbol{\hat{F}} = PostFusion(\boldsymbol{F}), \\
    &\boldsymbol{\hat{I}} = PixelShuffle(Conv(\boldsymbol{\hat{F}})).
\end{align}
The addition operation of the separate convolution of the output features is mathematically equivalent to $\boldsymbol{F} = Conv (Concat(\boldsymbol{\hat{F}_l},\boldsymbol{\hat{F}_h}))$, only doubles the number of bias parameters.  The $PostFusion$ module is an N$_f$ layer's transformer block, which will be detailed in Sec.~\ref{sec:sa-ca}.

\subsection{Dual-Branch Design and the SA-CA Module}
\label{sec:sa-ca}
After decomposing the moiré image $\boldsymbol{I_m}$ into $\boldsymbol{I_l}$ and $\boldsymbol{I_h}$ different parts, some awesome benefits are shown for later processing:
\vspace{-2mm}
\begin{enumerate}[left=0.5em, itemsep=-1pt]
    \item The high-frequency component shows strong spatial locality, carrying little color and mainly encoding texture. Thus, only local context is needed for effective demoiré, making the global image structure less important. This allows robust training via random cropping and direct application to ultra-high-resolution demoiré tasks.
    \item The low-frequency component is smooth and highly scale-robust. It can be downsampled to a much lower resolution (\textit{e.g.}, 0.1×) and still be accurately reconstructed. We experimented with $\boldsymbol{I_h}$ and $\boldsymbol{I_l}$ separately resized, and then restored to their original size to assess information loss. Results show reconstruction PSNRs of $\sim$25 dB ($\boldsymbol{I_h}$) and $\sim$50 dB ($\boldsymbol{I_l}$), confirming that $\boldsymbol{I_l}$ can be safely resized with minimal degradation, avoiding the risks of detail loss caused by downsampling~\cite{he2020fhde}.
    \item After resizing, moiré-induced color distortions become spatially concentrated. Consequently, only small, textureless patches with color deviations require correction, which still depends on local spatial features and channel-wise representations. Hence, using similar network structures for the high- and low-frequency branches is a natural and simplified choice.
    \item Aggressively downsampling the low-frequency component simplifies color shift correction, as full color information is preserved compactly. This lets the low-frequency branch use the same small-scale input during both training and inference phases, avoiding potential issues with varying input sizes.
\end{enumerate}

As analyzed, we adopt a two-branch learning framework and apply cropping and resizing strategies to the high- and low-frequency branches, respectively, shown in Fig.~\ref{fig:pipe_p1}. During inference, the high-frequency branch operates on the full-resolution input, while the low-frequency branch processes a downsampled image that is later upsampled back to full resolution. The resulting full-resolution features from both branches are then fused by the FCT module.

\begin{figure*}[t]
    \centering
    \includegraphics[width=0.98\linewidth]{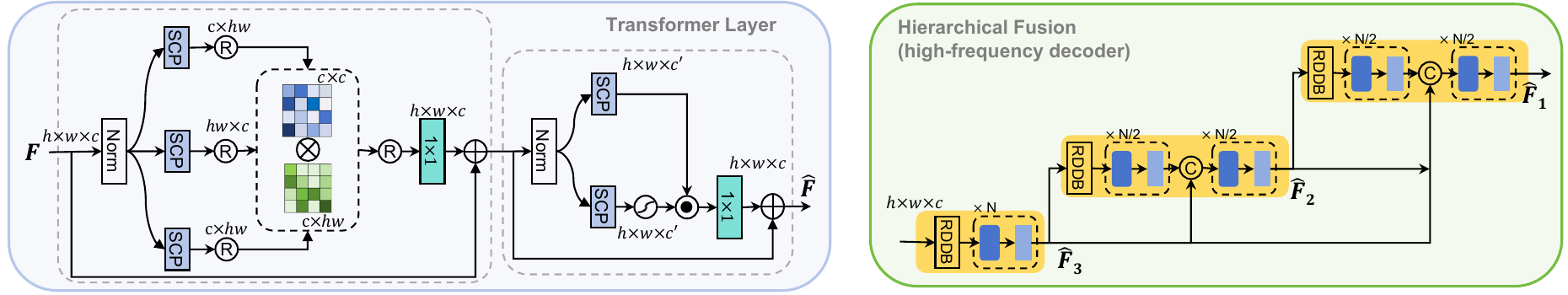}
    \vspace{-2mm}
    \caption{Left part is the detailed transformer architecture of the SA-CA module utilized in all the encoders and decoders. The right part shows the hierarchical fusion of the high-frequency branch.}
    \vspace{-5mm}
    \label{fig:pipe_p2}
\end{figure*}

\noindent \textbf{Spatial-Aware Channel Attention Module.} Inspired by the work~\cite{xiao2024p}, which reveals that the green channel is less affected by the moiré patterns, and according to Fig.~\ref{fig:first_comp}, we find that the RGB curves are not the same and sometimes show complementary effects, we decide to let each channel fully attend to others and adjust the feature. We adopt the channel attention mechanism here, which is also commonly used in~\cite{chen2023dual, restormer, xu2024image}. However, channel information alone is insufficient for precise image demoiréing—spatial information from the adjacent domain remains crucial for texture learning and color projection. This is because the frequency characteristics or directional patterns of a wave can only be inferred from multiple spatially adjacent points, instead of a single isolated point. We adopt the Residual Dilated Dense Block (RDDB)~\cite{zhang2018residual, huang2017densely, he2016deep, yu2015multi} for feature extraction with a large receptive field. Additionally, projection layers in the attention and FFN modules are crucial for aggregating spatial and channel features for local-feature and inter-channel learning.

Each block of the encoders and decoders uses the Spatial-Aware Channel Attention (SA-CA) Module. As shown in Fig.~\ref{fig:pipe_p2}, our whole Spatial-Aware Channel Attention Module is defined as follows: Given an input feature $\boldsymbol{F} \in \mathcal{R}^{h\times w\times c}$, it first goes through an RDDB block to extract the spatial information with a large receptive field. Then, it goes through N layers of transformer block, each consisting of a channel attention and a gated Feed-Forward Network.  For the channel Attention, given the input $\boldsymbol{F}$, we first generate the $\boldsymbol{Q, K, V} \in \mathcal{R}^{h\times w \times c}$ using the spatial-channel projection (SCP). Considering efficiency and functionality, we finally utilize the 1$\times$1 convolution followed by 3$\times$3 depth-wise convolution, expecting the former to learn the channel pattern projection and the latter to aggregate the adjacent information. Then using reshape and transpose, we get $\boldsymbol{Q^\prime} \in \mathcal{R}^{c\times hw}$, $\boldsymbol{K^\prime} \in \mathcal{R}^{hw\times c}$ and $\boldsymbol{V^\prime} \in \mathcal{R}^{c\times hw}$. And then, we perform channel attention $\boldsymbol{\hat{F}}= Attention(\boldsymbol{Q'}, \boldsymbol{K'}, \boldsymbol{V'}) + \boldsymbol{F}$, with channels divided into heads to learn different features. As for the subsequent Feed-Forward Network, the traditional one can only conduct channel-wise learning and hardly handle the spatial-channel variant demoiréing task. As a result, we decide to let the FFN be a gated module that filters the inappropriate information. Firstly, through the spatial-channel projection (also a 1$\times$1 convolution and a $3\times$3 depth-wise convolution), we get $\boldsymbol{X}, \boldsymbol{Y} \in \mathcal{R}^{h\times w \times c'}$. We utilize a non-linear function to get the filter scores and conduct the element-wise multiplication, followed by a convolution and a residual connection from the input $\boldsymbol{F}$.

Apart from the block design, we also design the high-frequency decoder's hierarchical fusion. As sometimes the moiré texture may have a wide range of variety, the demoiréed feature from the lower level of the decoder might help to produce the higher clean feature. We denote the output of each decoder block as $\boldsymbol{\hat{F_3}}, \boldsymbol{\hat{F_2}}, \boldsymbol{\hat{F_1}}$. For the mid-block of the decoder, the $\boldsymbol{\hat{F_3}}$ from the bottom block is concatenated into the medium output of the N/2 layers and then fed to the later N/2 layers for fusion. For the top block of the decoder, the  $\boldsymbol{\hat{F_3}}$ and $\boldsymbol{\hat{F_2}}$ are concatenated into the middle output of the N/2 layers and then fed to the later N/2 layers for fusion.
\vspace{-4mm}
\subsection{Training Strategy and Loss Functions}
\vspace{-2mm}
\label{sec:loss}
The training is set in two stages. For stage one, the high-frequency branch and the low-frequency branch are trained separately, with the crop/resize strategy, respectively. And for stage two, we jointly trained the two branches and the learnable FCT. It is noteworthy that, due to the different crop/resize strategy, we first interpolate the $\boldsymbol{\hat{F_l}}$ to the original full size. Then based on the crop location $(x_1, y_1, x_2, y_2)$ of $\boldsymbol{\hat{I_h}}$,  we crop the interpolated feature $\boldsymbol{\hat{F'_l}}$ and feed both $\boldsymbol{\hat{F'_l}}$ and $\boldsymbol{\hat{F_h}}$ into the learnable FCT to get the final $\boldsymbol{\hat{I}}$.

For the loss function of the first stage, we take the deep supervision strategy proposed in~\cite{zheng2020image}. The high- and low-branch loss function is defined as:
{\small
\begin{align}
    L^{high}_{stage_1} = \sum_{i=1}^{3} (L_1(\boldsymbol{\hat{I}_{h_i}}, \boldsymbol{I^{GT}_{h_i}}) + \lambda_1 \times L_p(\boldsymbol{\hat{I}_{h_i}}, \boldsymbol{I^{GT}_{h_i}}) ). \\
    L^{low}_{stage_1} = \sum_{i=1}^{3} (L_1(\boldsymbol{\hat{I}_{l_i}}, \boldsymbol{I^{GT}_{l_i}}) + \lambda_1 \times L_p(\boldsymbol{\hat{I}_{l_i}}, \boldsymbol{I^{GT}_{l_i}}) ).
\end{align} 
}
For the second stage, the loss function is defined as:
{\small
\begin{equation}
    L_{stage_2} = L_1(\boldsymbol{\hat{I}}, \boldsymbol{I^{GT}}) + \lambda_2 \times L_p(\boldsymbol{\hat{I}}, \boldsymbol{I^{GT}}) .
\end{equation}
}
We use the perceptual loss from the pretrained VGG16~\cite{simonyan2014very} Network. $\lambda_1$ and $\lambda_2$ are utilized to balance the two losses.

\vspace{-4mm}
\section{Experiments}
\vspace{-2mm}
\paragraph{Datasets and metrics.} In the experiments, we separately train and test our Freqformer on TIP2018~\cite{sun2018moire}, LCDMoire~\cite{zheng2020image}, FHDMi~\cite{he2020fhde} and UHDM~\cite{yu2022towards} datasets. TIP2018 contains 150,000 image pairs, with 135,000 used for training and 15,000 for testing. LCDMoire provides 10,000 images for training and 100 for validation. FHDMi consists of 9,981 and 2,019 image pairs for training and testing, respectively, at a resolution of $1920\times1080$. UHDM includes 4,500 and 500 4K image pairs for training and testing, respectively, and we follow~\cite{yu2022towards} by evaluating at $3840\times2160$ resolution. To measure the performance of all models, we use PSNR, SSIM~\cite{wang2004image} and LPIPS~\cite{zhang2018unreasonable} as the metrics.
Higher PSNR and SSIM, as well as lower LPIPS, indicate higher quality of restored images. PSNR is more sensitive to color variations, SSIM better captures structural moiré patterns, and LPIPS aligns more closely with human perceptual quality~\cite{he2020fhde}.

\paragraph{Implementation details.}  We implement our algorithm using Pytorch on an NVIDIA RTX 4090 GPU. For training on the TIP2018/LCDMoire/FHDMi/UHDM, we set the crop/resize size as 256/256, 512/512, 512/512 and 768/512, respectively. Following the settings of~\cite{yu2022towards}, we adopt the Adam optimizer~\cite{kingma2014adam} and the cyclic cosine schedule~\cite{loshchilov2016sgdr}, 150 epochs for both the training stages, and the batch size of 2. We set the $\lambda_1=1$ and $\lambda_2=0.1$. For the high-branch of the Freqformer, from the highest to the lowest resolution, the encoder's N is set to [2,2,2] and the decoder's N is set to [4,4,2]. For low-branch, both the encoder and decoder's N is [1,2,2]. N$_f$ of the learnable FCT is set to 2.

\begin{table*}[!t]

\centering
\resizebox{\textwidth}{!}{
\begin{tabular}{c|c|ccccccccc|c}
\toprule[1.2pt]
Dataset &Metrics   &Input &DMCNN~\cite{sun2018moire} &MDDM~\cite{MDDM} &WDNet~\cite{liu2020wavelet} &MopNet~\cite{he2019mop}  &MBCNN~\cite{zheng2020image} &FHDe$^{2}$Net~\cite{he2020fhde}   & ESDNet~\cite{yu2022towards} & ESDNet-L~\cite{yu2022towards} & Freqformer\\

\hline \multirow{2}{*}{TIP2018~\cite{sun2018moire}} & \text{PSNR}$\uparrow$ & 20.30 & 26.77 & N/A & 28.08 & 27.75 & 30.03 & 27.78 & 29.81 & \textcolor{blue}{30.11} & \textcolor{red}{30.63} \\
&\text{SSIM}$\uparrow$ & 0.7380 & 0.8710 & N/A & 0.9040 & 0.8950 & 0.8930 & 0.8960 & 0.9160 & \textcolor{blue}{0.9200} & \textcolor{red}{0.9269} \\

\hline \multirow{2}{*}{LCDMoire~\cite{zheng2020image}} &\text{PSNR}$\uparrow$  & 10.44 & 35.48 & 42.49 & 29.66 & N/A & 44.04 & 41.40 & 44.83 & \textcolor{blue}{45.34} & \textcolor{red}{45.62} \\
&\text{SSIM}$\uparrow$ &  0.5717 & 0.9785 & 0.9940 & 0.9670 & N/A & 0.9948 & N/A & 0.9963 & \textcolor{blue}{0.9966} & \textcolor{red}{0.9967} \\

\hline \multirow{3}{*}{FHDMi~\cite{he2020fhde}}
&\text{PSNR}$\uparrow$ &17.97  & 21.54 & 20.83 &N/A & 22.76 & 22.31 & 22.93  & 24.50 & \textcolor{blue}{24.88} & \textcolor{red}{25.26} \\

&\text{SSIM}$\uparrow$ & 0.7033 & 0.7727 & 0.7343 &N/A & 0.7958 & 0.8095 & 0.7885  & 0.8351 & \textcolor{blue}{0.8440} & \textcolor{red}{0.8518} \\

&\text{LPIPS}$\downarrow$ &0.2837 & 0.2477 & 0.2515 &N/A & 0.1794 & 0.1980 & 0.1688  & 0.1354 & \textcolor{blue}{0.1301} & \textcolor{red}{0.1253} \\

\hline \multirow{3}{*}{UHDM~\cite{yu2022towards}}
&\text{PSNR}$\uparrow$ &17.12 &19.91 &20.09 &20.36 &19.49 &21.41 &20.34 &22.12 &\textcolor{red}{22.42} & \textcolor{blue}{22.24}\\

&\text{SSIM}$\uparrow$ &0.5089 &0.7575 &0.7441 &0.6497 &0.7572 &0.7932 &0.7496 &0.7956 & \textcolor{blue}{0.7985} & \textcolor{red}{0.8021} \\

&\text{LPIPS}$\downarrow$ &0.5314 &0.3764 &0.3409 &0.4882 &0.3857 &0.3318 &0.3519 &0.2551 &\textcolor{blue}{0.2454} & \textcolor{red}{0.2424} \\

\hline
\hline

- & Params (M) & - & 1.426 & 7.637 & 3.360 & 58.565 & 14.192 & 13.571 & 5.934 & 10.623 & 6.065 \\

\bottomrule[1.2pt]
\end{tabular}
}
\caption{Quantitative comparisons with sRGB image demoiréing methods. \textcolor{red}{Red}: best and \textcolor{blue}{Blue}: second-best.}
\label{tab: main_tab}
\vspace{-3.5mm}
\end{table*}

\begin{table*}[!t]
\centering
\resizebox{\textwidth}{!}{
\begin{tabular}{c|c|ccccc|ccc|c}
\toprule[1.2pt]
Dataset &Metrics &OSEDiff~\cite{wu2024one} &AdaIR*~\cite{cui2025adair} & AdaIR~\cite{cui2025adair}  &MoCE-IR*~\cite{zamfir2024complexityexperts} &MoCE-IR~\cite{zamfir2024complexityexperts} & RRID~\cite{xu2024image} & STD-Net~\cite{STDNet} & DTNet~\cite{xu2024direction} & Freqformer\\

\hline \multirow{3}{*}{FHDMi~\cite{he2020fhde}}
&\text{PSNR}$\uparrow$ & 21.81  & 17.08 & 21.87 & 17.40 & 21.80 & 24.39 & \textcolor{blue}{25.05} & 23.01 & \textcolor{red}{25.26} \\

&\text{SSIM}$\uparrow$ & 0.7764 & 0.6986 & 0.8041 & 0.7031 & 0.7943 & 0.8300 & \textcolor{blue}{0.8414} & 0.8128 & \textcolor{red}{0.8518}\\

&\text{LPIPS}$\downarrow$ & 0.1516 & 0.3057 & 0.1924 & 0.2969 & 0.2026 & N/A & \textcolor{blue}{0.1293} & 0.1548 & \textcolor{red}{0.1253} \\

\hline \multirow{3}{*}{UHDM~\cite{yu2022towards}}
&\text{PSNR}$\uparrow$ & 20.66 & 16.68 & 20.96 & 16.97 & 20.46 & \textcolor{blue}{21.98} & N/A & 19.74 & \textcolor{red}{22.24} \\

&\text{SSIM}$\uparrow$ & 0.7607 & 0.4949 & 0.7807 & 0.4971 & 0.7644 & \textcolor{blue}{0.7935} & N/A & 0.7637 & \textcolor{red}{0.8021} \\

&\text{LPIPS}$\downarrow$ & \textcolor{red}{0.2394} & 0.5440 & 0.3040 & 0.5366 & 0.3249 & 0.3032 & N/A & 0.2624 & \textcolor{blue}{0.2424} \\

\hline
\hline

- & Params (M) & 950 & 28.785 & 28.785 & 11.478 & 11.478 & 1.448 & 27.960 & 7.360 & 6.065 \\

\bottomrule[1.2pt]
\end{tabular}}
\caption{Quantitative comparison with image restoration models and other types of demoiréing models. \textcolor{red}{Red}: best and \textcolor{blue}{Blue}: second-best.}
\label{tab: comparison}
\vspace{-4.5mm}
\end{table*}

\vspace{-1mm}
\subsection{Comparisons with State-of-the-Art Methods}
\vspace{-1mm}

\paragraph{Quantitative comparison.}
In Table~\ref{tab: main_tab}, our proposed Freqformer achieves state-of-the-art performance compared with other image demoiréing methods, including DMCNN~\cite{sun2018moire}, MDDM~\cite{MDDM}, WDNet~\cite{liu2020wavelet}, MopNet~\cite{he2019mop}, MBCNN~\cite{zheng2020image}, FHDe$^{2}$Net~\cite{he2020fhde}, and ESDNet~\cite{yu2022towards}.
On TIP2018/LCDMoire/FHDMi datasets, Freqformer significantly outperforms other methods by a large margin in all evaluation metrics. This indicates superior reconstruction quality, structural fidelity, and perceptual similarity.
On the more challenging UHDM dataset, our Freqformer ranks first in SSIM and LPIPS, and second in PSNR, closely following ESDNet-L. 
Notably, Freqformer achieves these with only 6.065M parameters, outperforming or matching methods with significantly larger model sizes (\textit{e.g.}, MopNet with 58.565M and MBCNN with 14.192M), highlighting its excellent trade-off between performance and model efficiency. 

Additionally, in Table~\ref{tab: comparison}, we compare our method with several recent general-purpose image restoration models, including AdaIR~\cite{cui2025adair} and MoCE-IR~\cite{zamfir2024complexityexperts}, as well as the generative restoration model OSEDiff~\cite{wu2024one}. The inclusion of OSEDiff is motivated by the recent success of generative priors in related image restoration tasks, like~\cite{wang2026strsr, liu2025osdd, liu2025fidediff, Guo_2025_ICCV}. We further compare our method with the state-of-the-art RAW-sRGB image demoiréing model RRID~\cite{xu2024image}, as well as the video demoiréing methods STD-Net~\cite{STDNet} and DTNet~\cite{xu2024direction}. The AdaIR$^*$ and MoCE-IR$^*$ denote direct testing with their all-in-one model, while without $^*$ represents the retraining on the two datasets. For models RRID, STD-Net, and DTNet, we adapt their network architectures according to their papers and retrain them on the datasets. Our model surpasses all these models, only inferior to OSEDiff in LPIPS, which has nearly 1B parameters.


\begin{figure*}[t]
\scriptsize
\centering
\begin{tabular}{c}
\begin{adjustbox}{valign=t}
\begin{tabular}{cccccccc}
{\includegraphics[width=0.2\textwidth,height=0.12\textwidth]{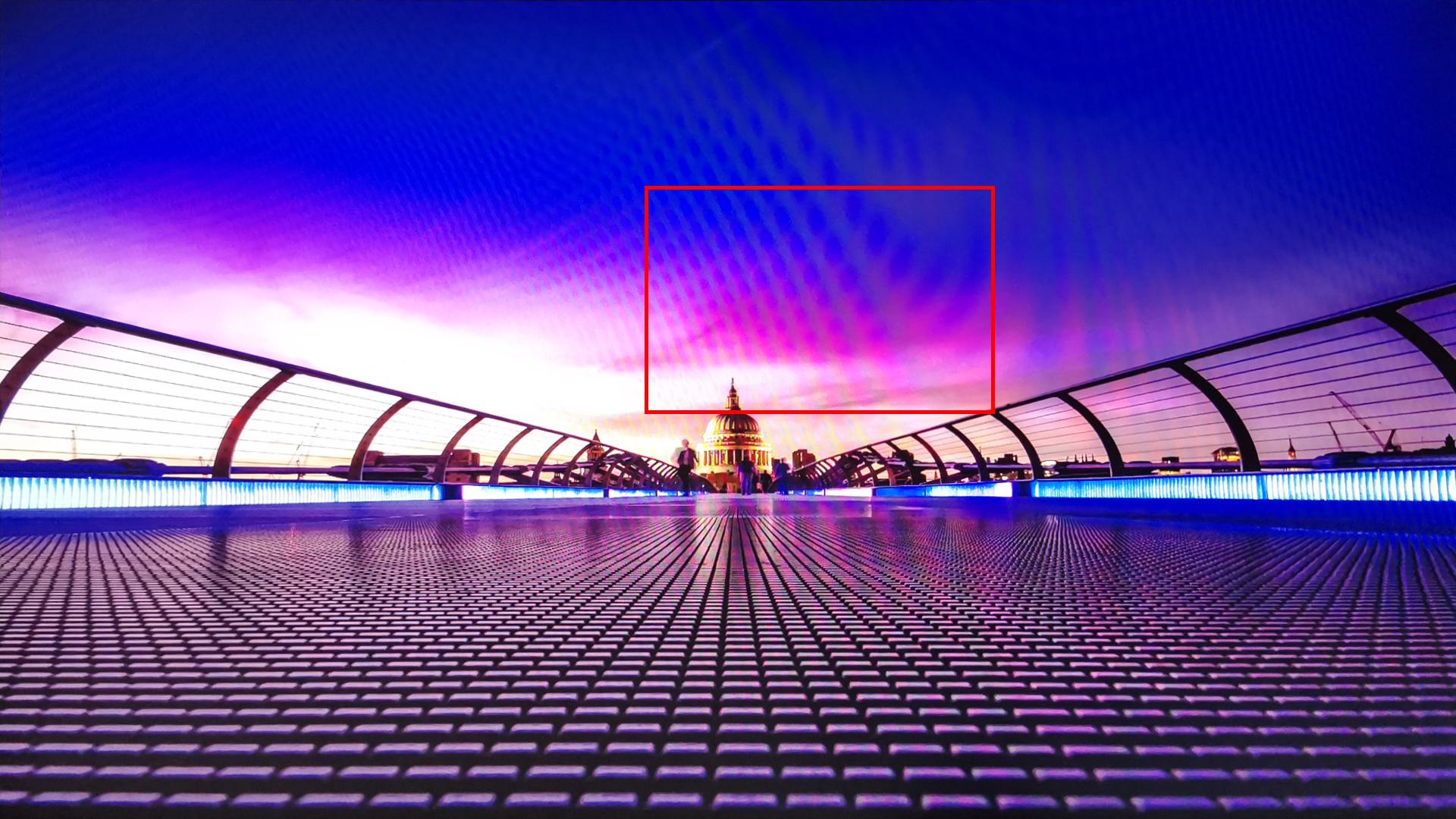}} \hspace{-2mm} &
\includegraphics[width=0.11\textwidth,height=0.12\textwidth]{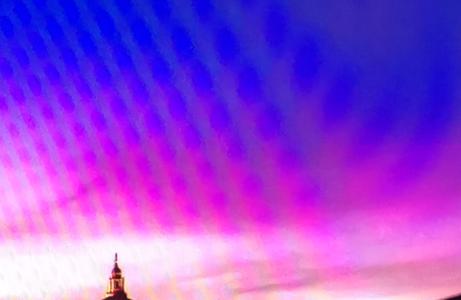} \hspace{-2mm} &
\includegraphics[width=0.11\textwidth,height=0.12\textwidth]{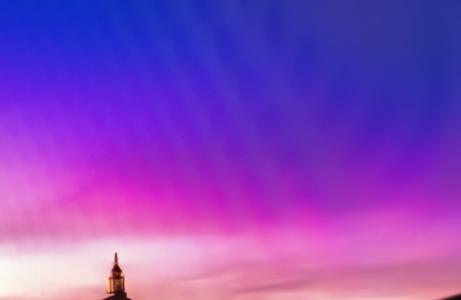} \hspace{-2mm} &
\includegraphics[width=0.11\textwidth,height=0.12\textwidth]{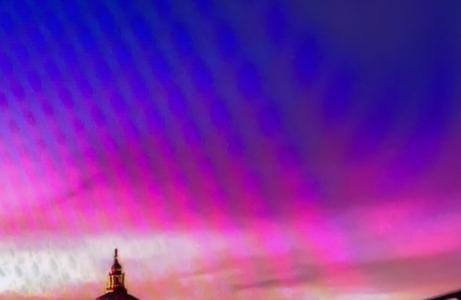} \hspace{-2mm} &
\includegraphics[width=0.11\textwidth,height=0.12\textwidth]{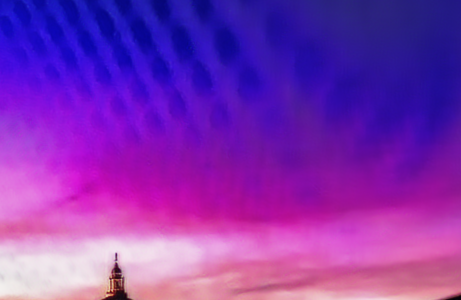} \hspace{-2mm} &
\includegraphics[width=0.11\textwidth,height=0.12\textwidth]{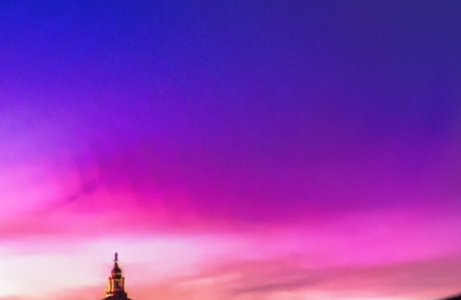} \hspace{-2mm} &
\includegraphics[width=0.11\textwidth,height=0.12\textwidth]{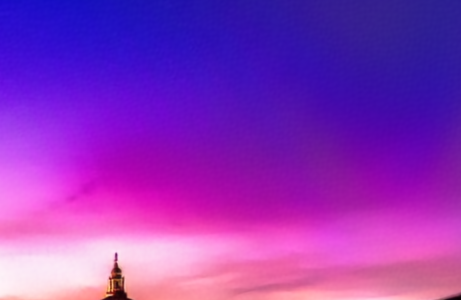} \hspace{-2mm} &
\includegraphics[width=0.11\textwidth,height=0.12\textwidth]{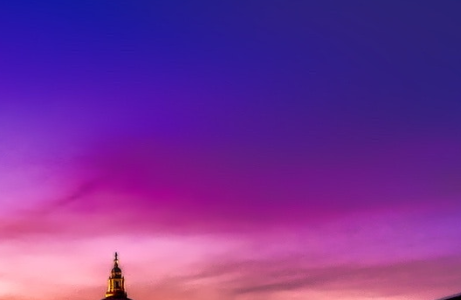} \hspace{-2mm} 
\\
\scalebox{0.7}{FHDMi-03279~\cite{he2020fhde}} \hspace{-2mm} &
\scalebox{0.7}{Input} \hspace{-2mm}  &
\scalebox{0.7}{OSEDiff~\cite{wu2024one}} \hspace{-2mm} &
\scalebox{0.7}{AdaIR~\cite{cui2025adair}} \hspace{-2mm} &
\scalebox{0.7}{FHDe$^2$Net~\cite{he2020fhde}} \hspace{-2mm}  &
\scalebox{0.7}{ESDNet-L~\cite{yu2022towards}} \hspace{-2mm} &
\scalebox{0.7}{Freqformer} \hspace{-2mm}  &
\scalebox{0.7}{GT} \hspace{-2mm} \\

{\includegraphics[width=0.2\textwidth,height=0.12\textwidth]{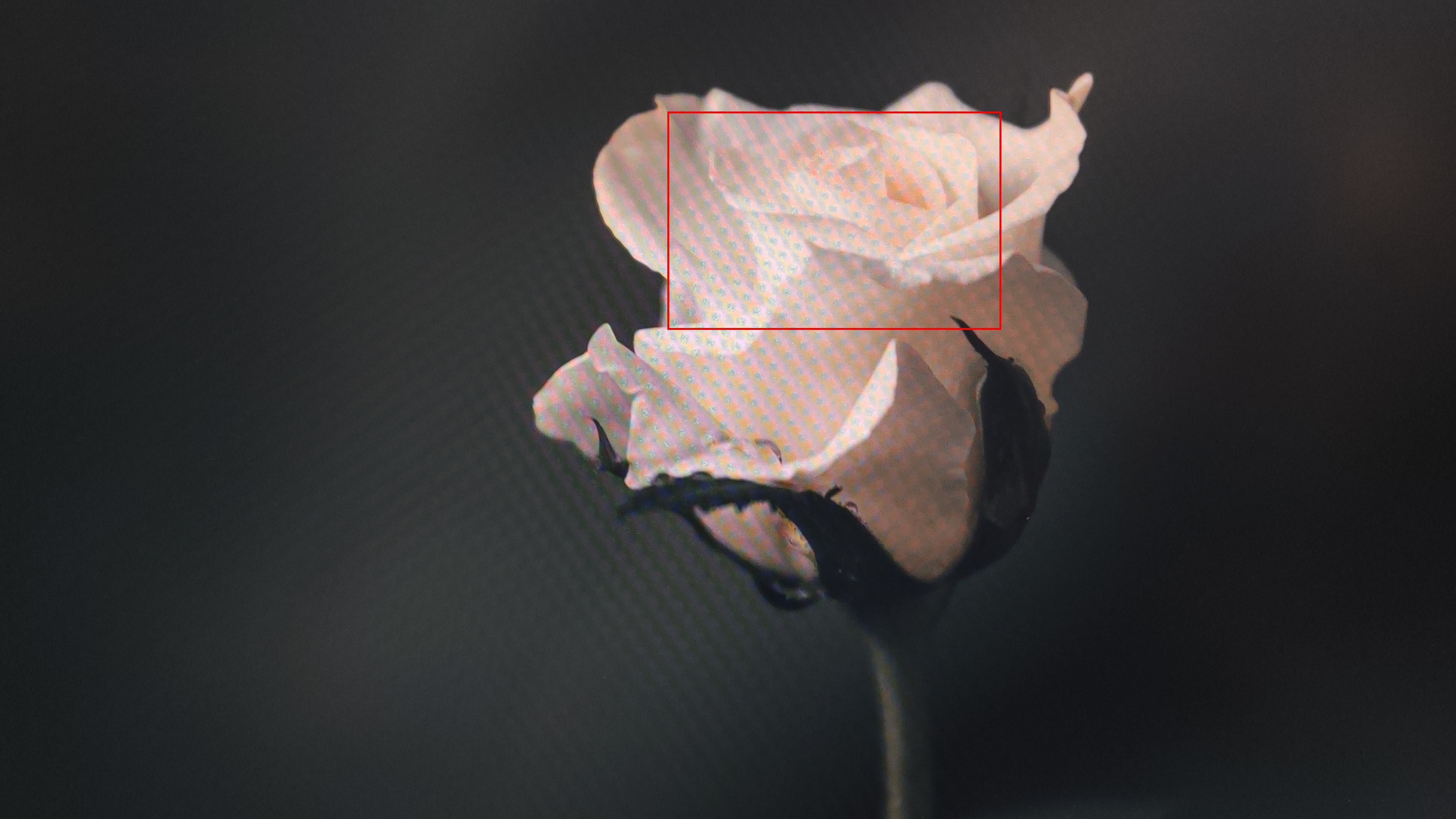}} \hspace{-2mm} &
\includegraphics[width=0.11\textwidth,height=0.12\textwidth]{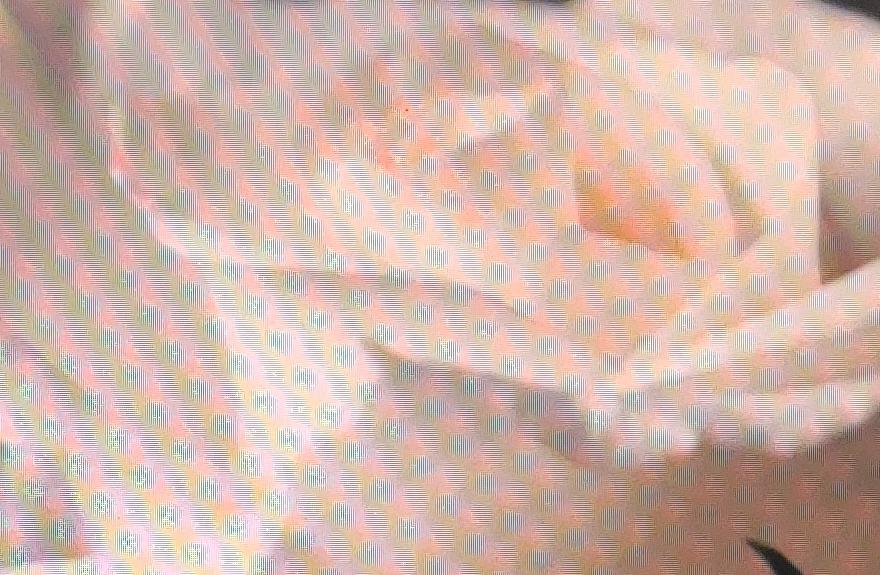} \hspace{-2mm} &
\includegraphics[width=0.11\textwidth,height=0.12\textwidth]{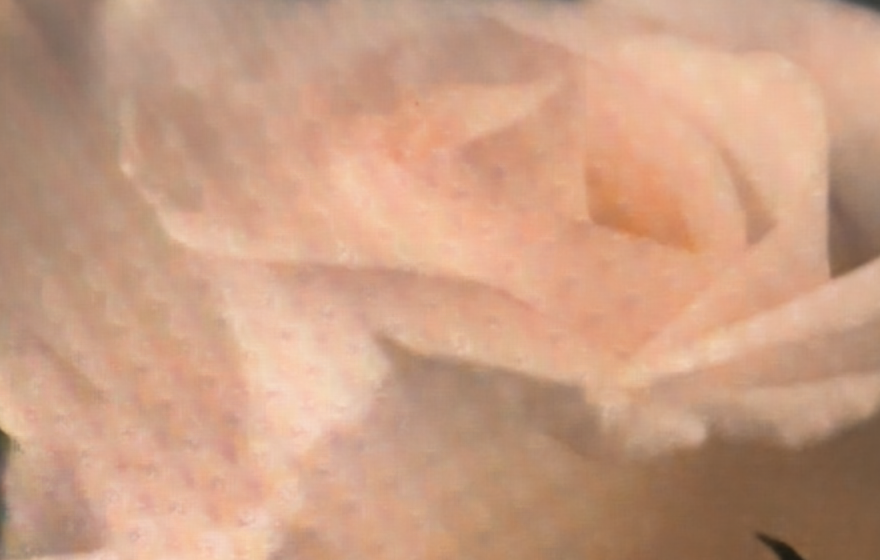} \hspace{-2mm} &
\includegraphics[width=0.11\textwidth,height=0.12\textwidth]{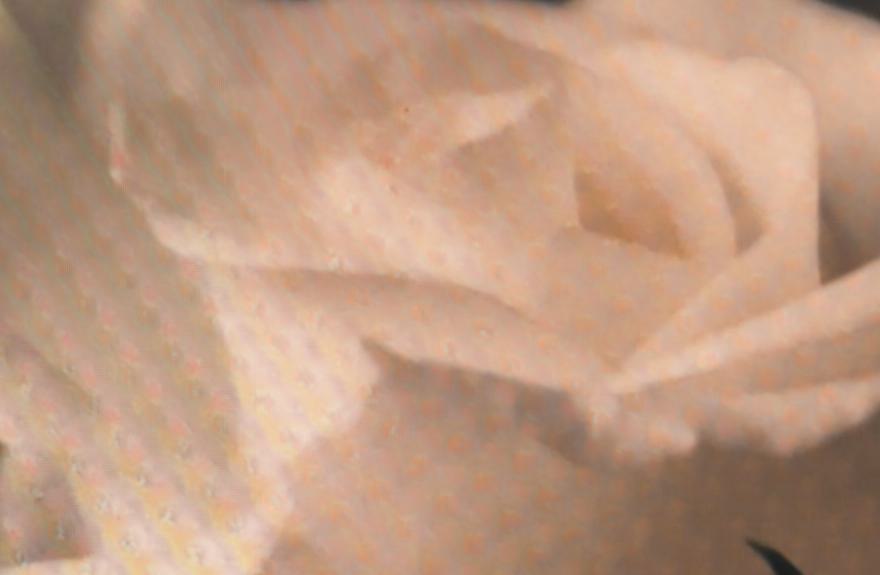} \hspace{-2mm} &
\includegraphics[width=0.11\textwidth,height=0.12\textwidth]{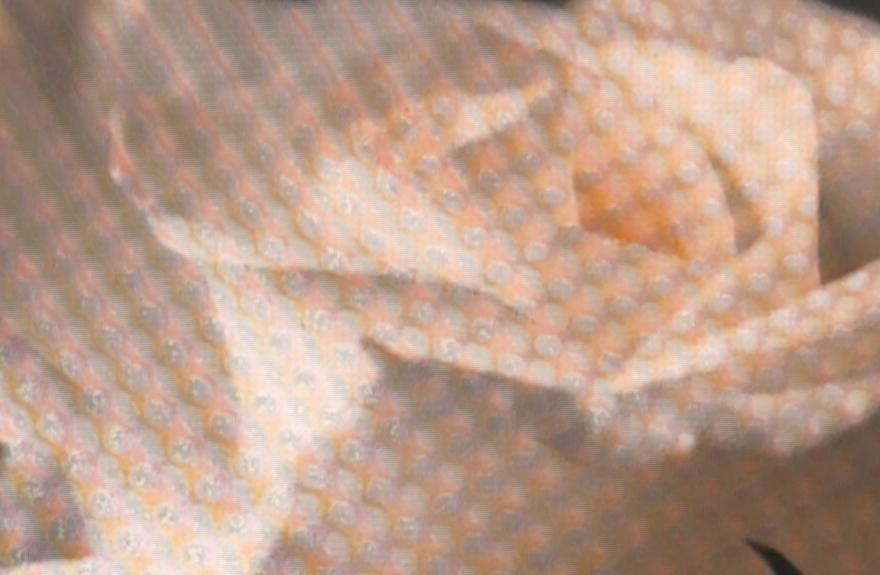} \hspace{-2mm} &
\includegraphics[width=0.11\textwidth,height=0.12\textwidth]{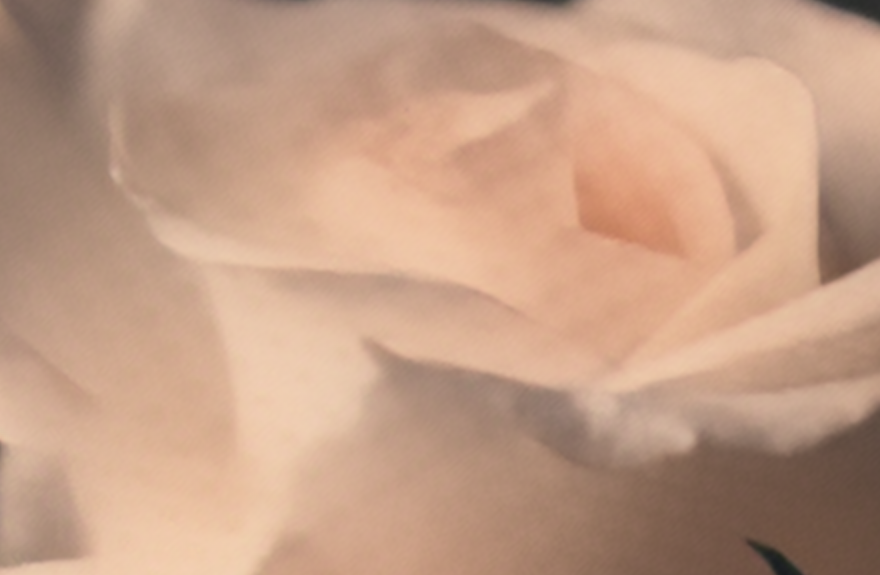} \hspace{-2mm} &
\includegraphics[width=0.11\textwidth,height=0.12\textwidth]{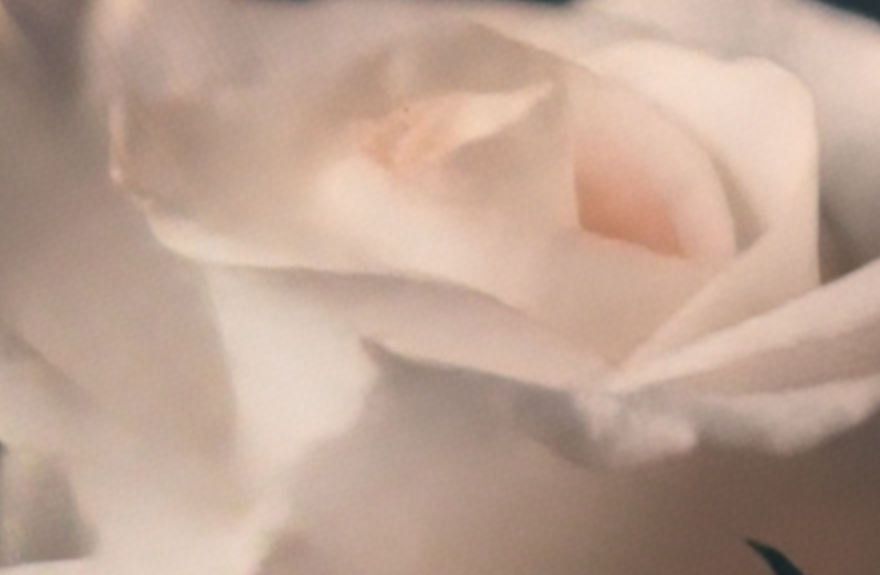} \hspace{-2mm} &
\includegraphics[width=0.11\textwidth,height=0.12\textwidth]{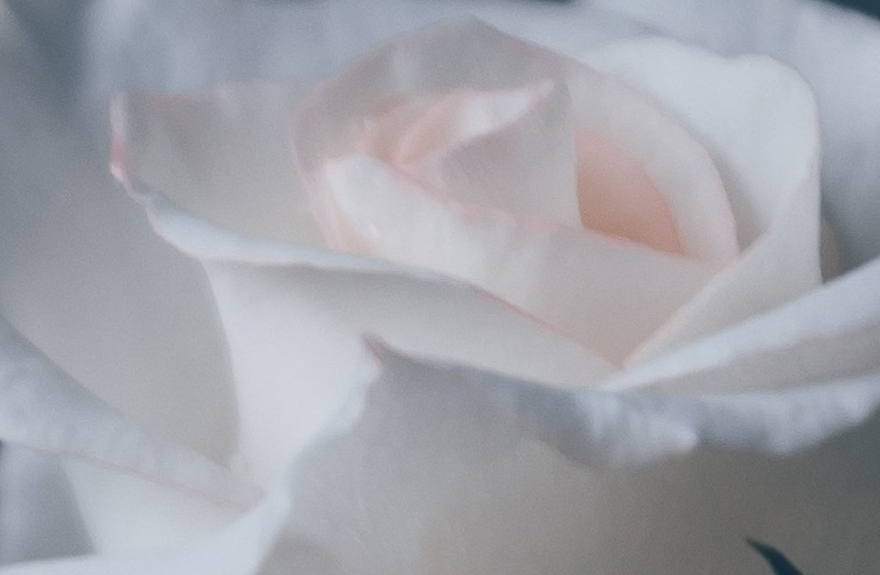} \hspace{-2mm} 

\\
\scalebox{0.7}{UHDM-0208~\cite{yu2022towards}} \hspace{-2mm} &
\scalebox{0.7}{Input} \hspace{-2mm}  &
\scalebox{0.7}{RRID~\cite{xu2024image}} \hspace{-2mm} &
\scalebox{0.7}{AdaIR~\cite{cui2025adair}} \hspace{-2mm} &
\scalebox{0.7}{MoCE-IR~\cite{zamfir2024complexityexperts}} \hspace{-2mm}  &
\scalebox{0.7}{ESDNet-L~\cite{yu2022towards}} \hspace{-2mm} &
\scalebox{0.7}{Freqformer} \hspace{-2mm}  &
\scalebox{0.7}{GT} \hspace{-2mm}
\\
{\includegraphics[width=0.2\textwidth,height=0.12\textwidth]{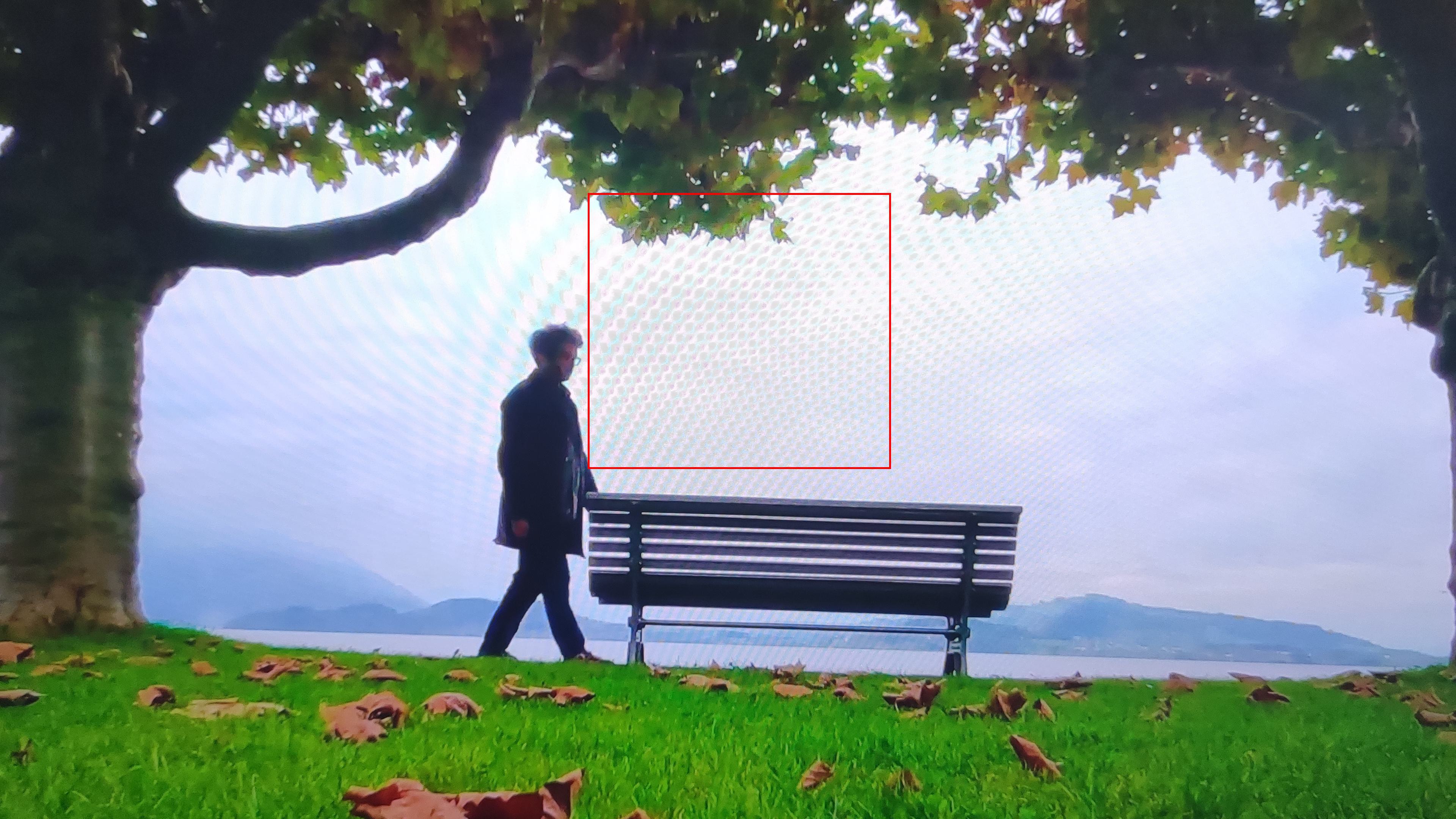}} \hspace{-2mm} &
\includegraphics[width=0.11\textwidth,height=0.12\textwidth]{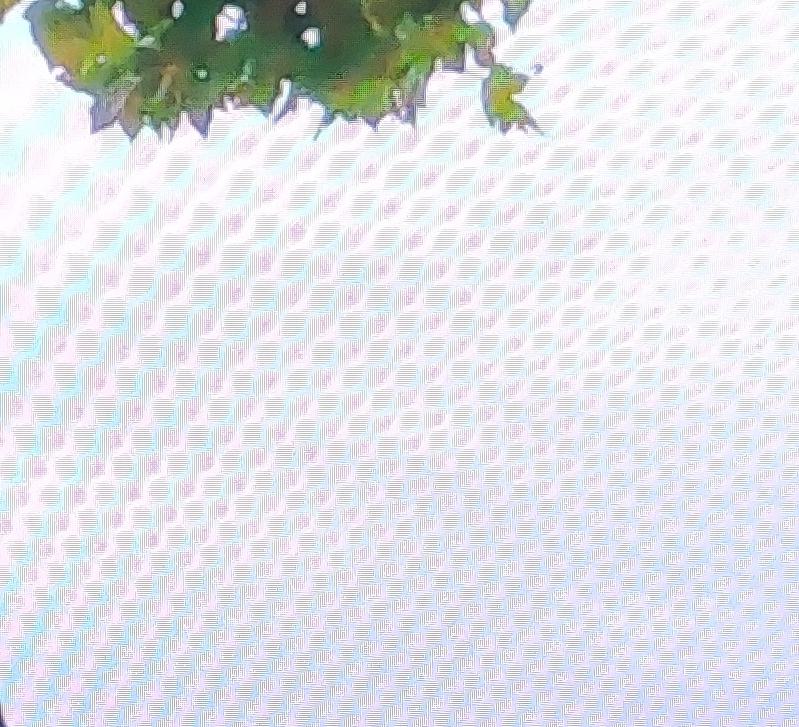} \hspace{-2mm} &
\includegraphics[width=0.11\textwidth,height=0.12\textwidth]{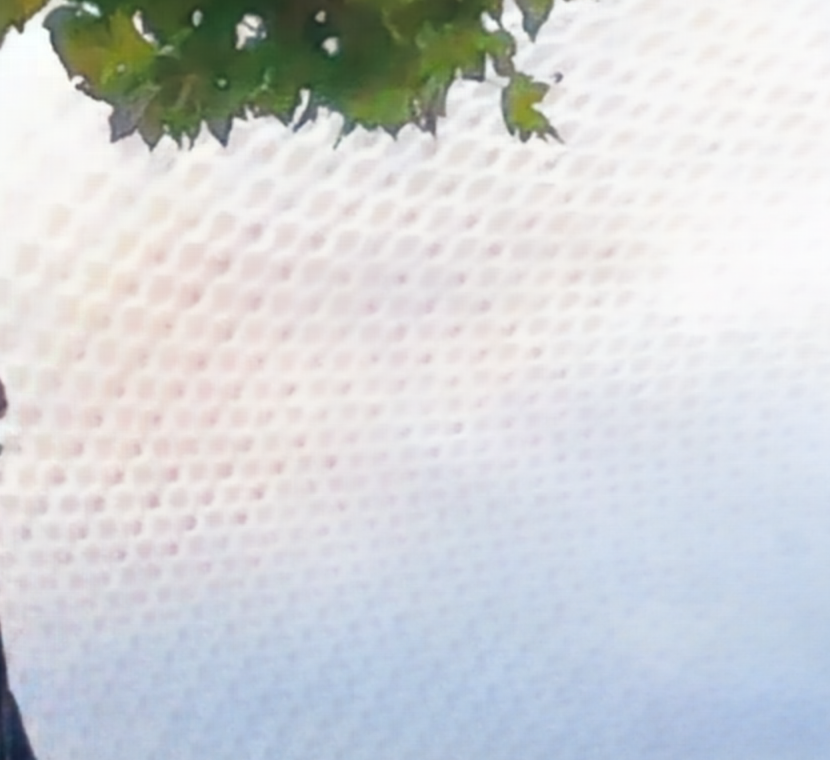} \hspace{-2mm} &
\includegraphics[width=0.11\textwidth,height=0.12\textwidth]{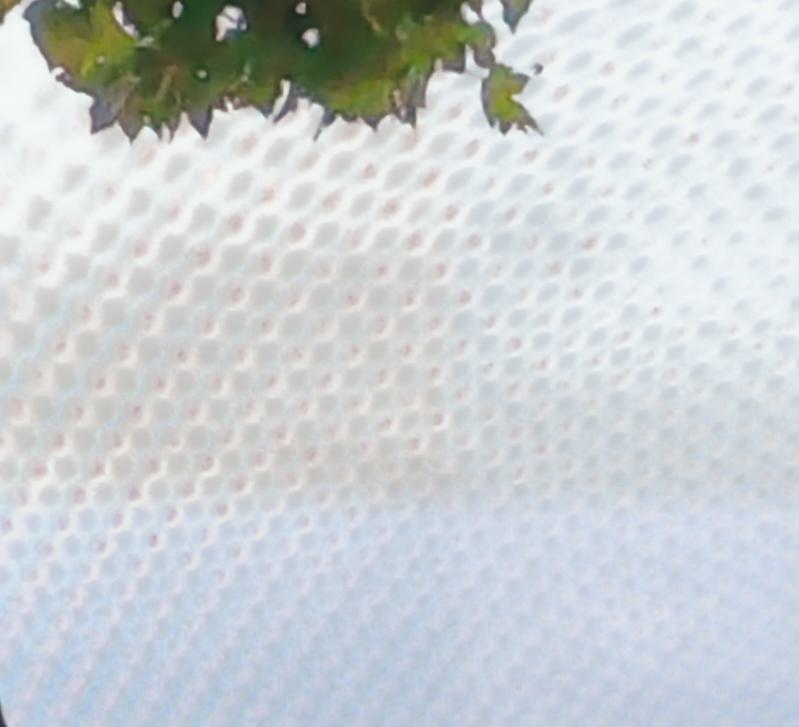} \hspace{-2mm} &
\includegraphics[width=0.11\textwidth,height=0.12\textwidth]{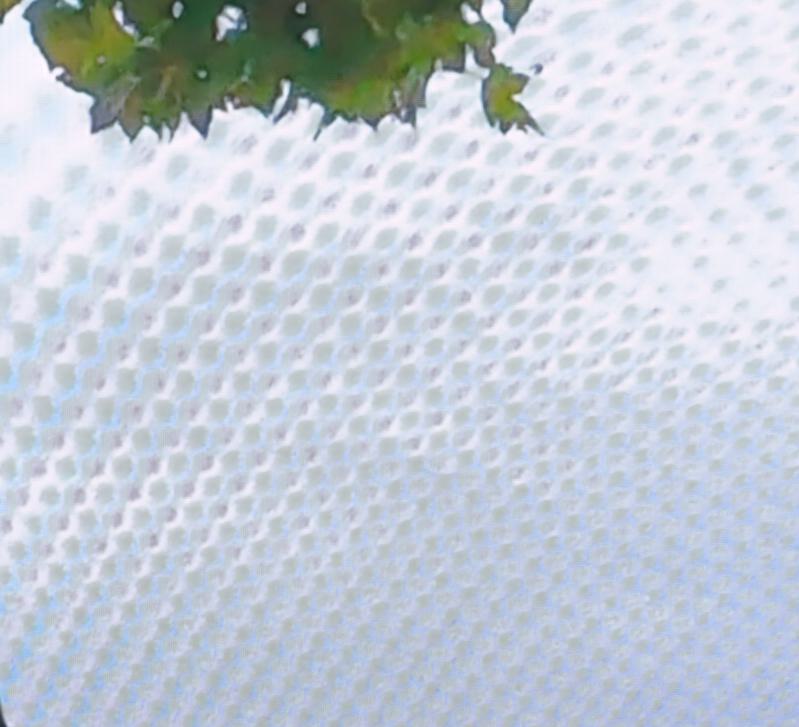} \hspace{-2mm} &
\includegraphics[width=0.11\textwidth,height=0.12\textwidth]{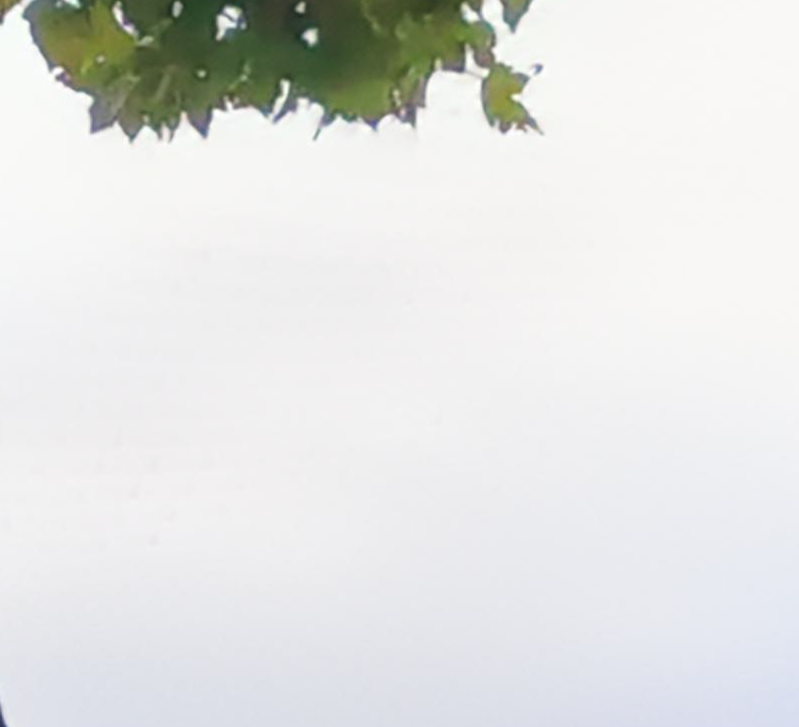} \hspace{-2mm} &
\includegraphics[width=0.11\textwidth,height=0.12\textwidth]{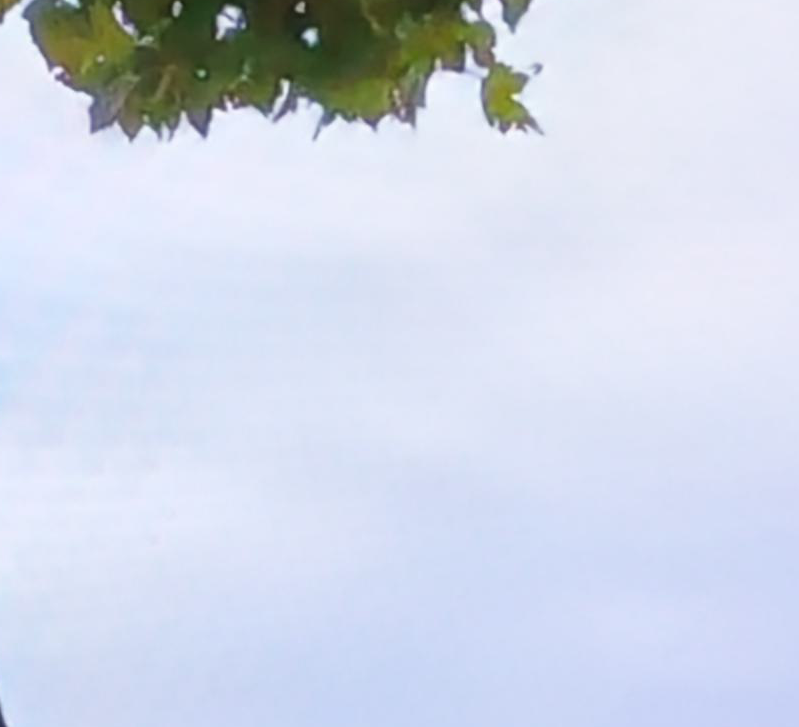} \hspace{-2mm} &
\includegraphics[width=0.11\textwidth,height=0.12\textwidth]{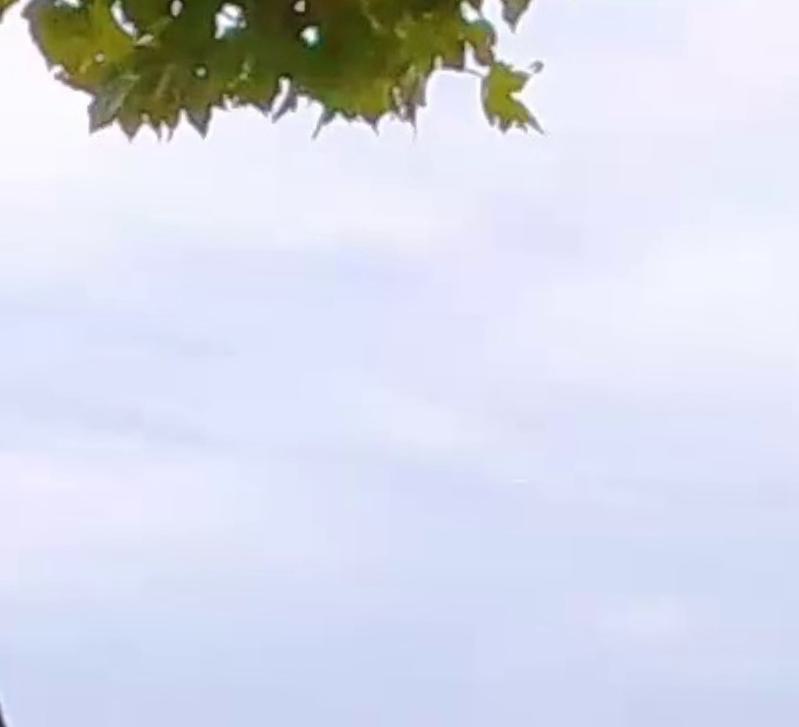} \hspace{-2mm} 
\\
\scalebox{0.7}{UHDM-0096~\cite{yu2022towards}} \hspace{-2mm} &
\scalebox{0.7}{Input} \hspace{-2mm}  &
\scalebox{0.7}{RRID~\cite{xu2024image}} \hspace{-2mm} &
\scalebox{0.7}{AdaIR~\cite{cui2025adair}} \hspace{-2mm} &
\scalebox{0.7}{MoCE-IR~\cite{zamfir2024complexityexperts}} \hspace{-2mm}  &
\scalebox{0.7}{ESDNet-L~\cite{yu2022towards}} \hspace{-2mm} &
\scalebox{0.7}{Freqformer} \hspace{-2mm}  &
\scalebox{0.7}{GT} \hspace{-2mm} 
\\

\end{tabular}
\end{adjustbox}

\end{tabular}
\vspace{-3mm}
\caption{Visual comparisons of different baselines on FHDMi and UHDM datasets (please zoom in for better details). While OSEDiff, AdaIR, MoCE-IR, and FHDe$^2$Net all struggle with removing moiré artifacts, ESDNet-L also leaves residual traces behind. In addition, these baselines suffer from color distortions and fail to accurately recover the original tones. In contrast, our approach effectively eliminates both moiré and color artifacts, as shown in the last row.}
\vspace{-5mm}
\label{fig:vis_comp}
\end{figure*}

\paragraph{Qualitative comparison.}
Figure~\ref{fig:vis_comp} presents visual comparisons of image restoration results on the FHDMi and UHDM datasets. 
Compared with existing methods, our Freqformer not only removes moiré more effectively, preserves fine details with high fidelity, but also mitigates the effects of color distortions and recovers the true color representation, resulting in more realistic and visually pleasing outputs. 
Competing methods such as OSEDiff, AdaIR, MoCE-IR, and RRID all struggle with moiré patterns that are embedded within the original content. Although ESDNet-L performs better than the others, it still leaves residual moiré artifacts, as observed in the top two rows of Fig.~\ref{fig:vis_comp}. Moreover, in the last row, ESDNet-L and all other baselines exhibit moiré-induced color distortions and fail to restore the original colors accurately. In contrast, our method successfully eliminates both the moiré patterns and color distortions,  demonstrating Freqformer's superior ability in both artifact removal and detail restoration, consistent with its leading performance in PSNR, SSIM, and LPIPS.

\paragraph{Model Complexity.} 
As compared in Tabs.~\ref{tab: main_tab} and~\ref{tab: comparison}, Freqformer not only has fewer parameters compared to MopNet, MBCNN, AdaIR and MoCE-IR, etc, but also achieves a smaller size than ESDNet-L, while superior to both ESDNet and ESDNet-L. Moreover, Freqformer requires only 2.46 TFLOPS to process 4K images, which is on par with ESDNet (2.24T) and lower than ESDNet-L (3.68T), AdaIR (31.06T) and MoCE-IR (5.43T). In addition, for 4K image inference, Freqformer achieves an inference speed of 0.58 s/image, which is slower than that of ESDNet-L (0.25 s/image). Freqformer requires fewer FLOPS but is slower, likely due to the standard attention operations being less optimized for operator fusion, kernel scheduling, and hardware acceleration than purely convolution-based models. This will be one of our future optimization directions.

\vspace{-2mm}
\subsection{Ablation Study}
\vspace{-1mm}
In this section, we conduct ablation studies on FHDMi to validate our frequency decomposition, training strategy, and network architecture.

\begin{figure}[h]
\scriptsize
\centering
\vspace{-9mm}
\begin{tabular}{c}
\hspace{-0.45cm}
\begin{adjustbox}{valign=t}
\begin{tabular}{cccccc}
\includegraphics[height=0.12\textwidth]{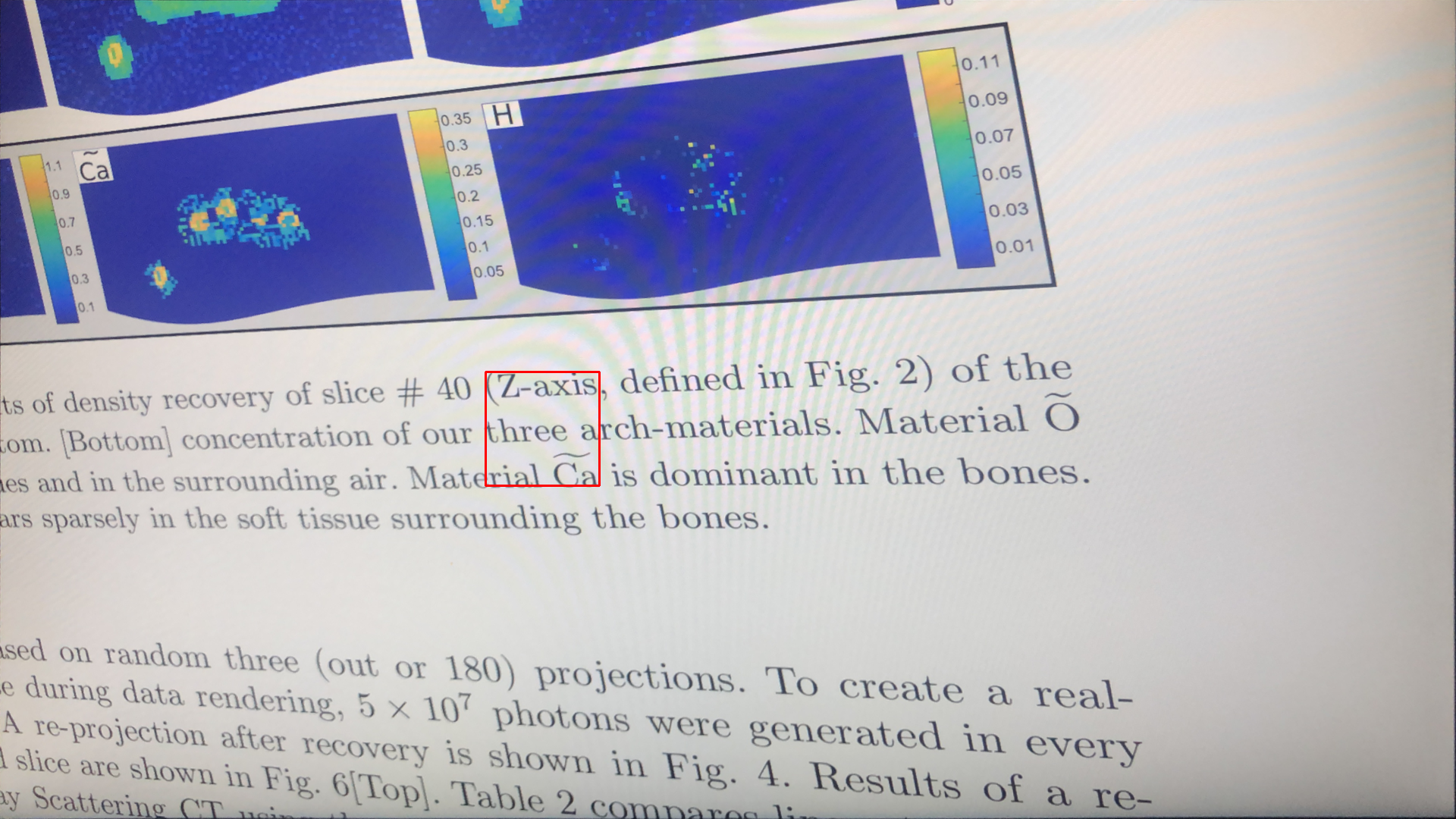} \hspace{-1mm} &
\includegraphics[width=0.12\textwidth,height=0.12\textwidth]{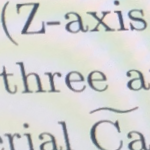} \hspace{-1mm} &
\includegraphics[width=0.12\textwidth,height=0.12\textwidth]{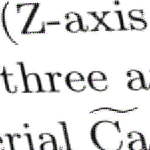} \hspace{-1mm} &
\includegraphics[width=0.12\textwidth,height=0.12\textwidth]{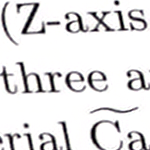} \hspace{-1mm} &
\includegraphics[width=0.12\textwidth,height=0.12\textwidth]{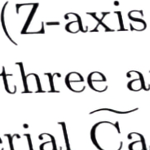} \hspace{-1mm} &
\includegraphics[width=0.12\textwidth,height=0.12\textwidth]{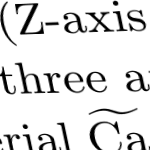} \hspace{-1mm} 
\\
FHDMi-00334 \hspace{-1mm} &
LQ \hspace{-1mm} &
Haar \hspace{-1mm}  &
DCT \hspace{-1mm}  &
Ours \hspace{-1mm} &
GT \hspace{-1mm} 
\\
\includegraphics[height=0.12\textwidth]{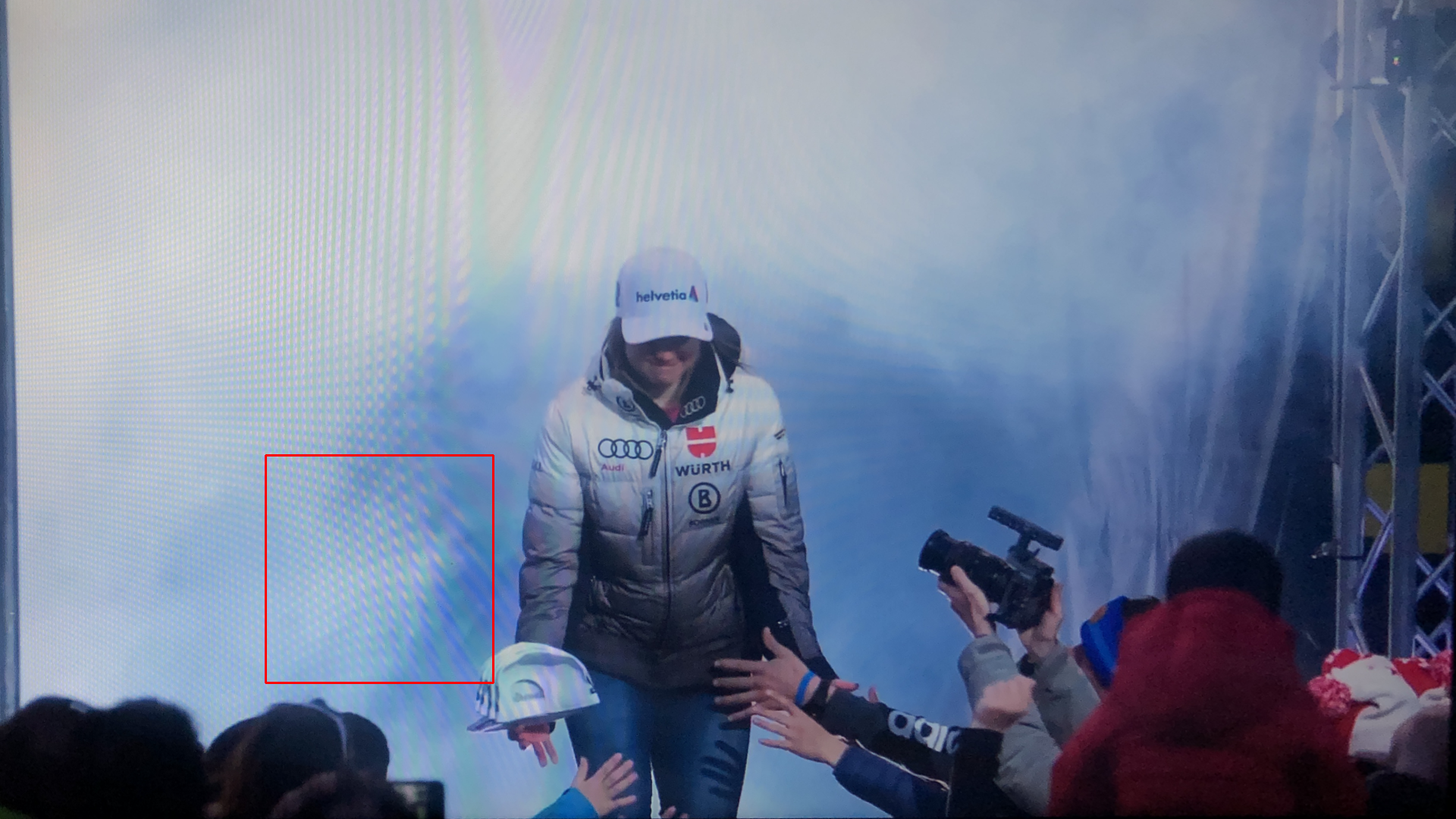} \hspace{-1mm} &
\includegraphics[width=0.12\textwidth,height=0.12\textwidth]{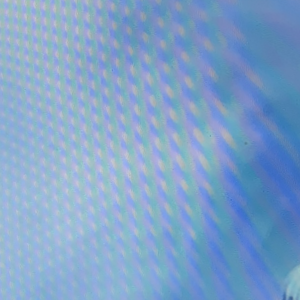} \hspace{-1mm} &
\includegraphics[width=0.12\textwidth,height=0.12\textwidth]{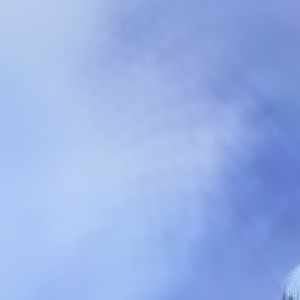} \hspace{-1mm} &
\includegraphics[width=0.12\textwidth,height=0.12\textwidth]{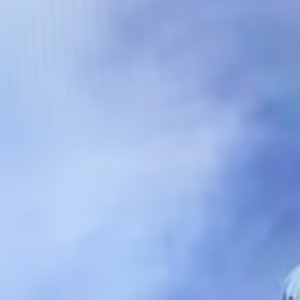} \hspace{-1mm} &
\includegraphics[width=0.12\textwidth,height=0.12\textwidth]{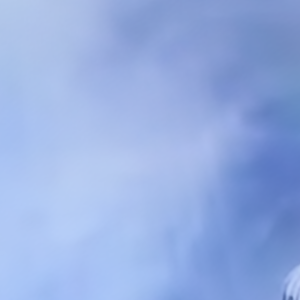} \hspace{-1mm} &
\includegraphics[width=0.12\textwidth,height=0.12\textwidth]{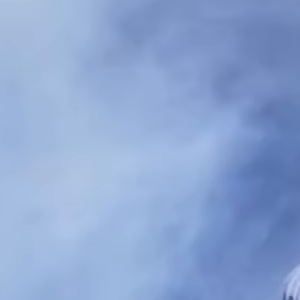} \hspace{-1mm} 
\\
FHDMi-00298 \hspace{-1mm} &
LQ \hspace{-1mm} &
Haar \hspace{-1mm}  &
DCT \hspace{-1mm}  &
Ours \hspace{-1mm} &
GT \hspace{-1mm} 
\\
\end{tabular}
\end{adjustbox}

\end{tabular}
\vspace{-2mm}
\caption{Visual comparisons of the Haar DWT, Block DCT, and ours.}
\vspace{-5mm}
\label{fig:aba_haar}
\end{figure}

\textbf{Our frequency decomposition v.s. Haar DWT and Block DCT.} We compare our approach against two alternative transforms: the 2-level Haar DWT~\cite{luowavelet2020cvprw, liu2020wavelet} (yielding a 10.921M-parameter model with band-specific demoiréing modules) and the Block DCT~\cite{zheng2020image, he2020fhde} with a block size of 8 (yielding a 12.715M-parameter model). As reported in Tab.~\ref{tab:haatdctcmp}, Freqformer equipped with our proposed transform significantly outperforms both variants. Furthermore, visual comparisons in Fig.~\ref{fig:aba_haar} demonstrate that images restored using either Haar DWT or Block DCT suffer from severe detail degradation. Specifically, the first row of Fig.~\ref{fig:aba_haar} shows that text edges appear distorted, jagged, and blurred. Moreover, as illustrated in the second row of Fig.~\ref{fig:aba_haar}, neither of these alternative transforms completely eradicates the moiré patterns, leaving visible residual artifacts.

\begin{table}[htbp]
    \centering
    \vspace{-3mm}
    \setlength{\tabcolsep}{4pt}
    \resizebox{0.7 \textwidth}{!}{
    \begin{tabular}{l|ccc|c}
        \toprule
        Model & PSNR & SSIM & LPIPS & Params (M) \\
        \hline
        Freqformer & 25.26 & 0.8518 & 0.1253 & 6.065 \\
        Freqformer w/ Haar DWT & 23.01 & 0.8036 & 0.1810 & 10.921 \\
        Freqformer w/ Block DCT & 23.22 & 0.8073 & 0.1569 & 12.715 \\
        \bottomrule
    \end{tabular}}
    \caption{Frequency Decomposition Comparison.}
    \vspace{-7mm}
    \label{tab:haatdctcmp}
\end{table}

\begin{table}[htbp]
    \centering
    \setlength{\tabcolsep}{4pt}
    \resizebox{0.7 \textwidth}{!}{
    \begin{tabular}{l|ccc|c}
        \toprule
        Model (Low branch) & PSNR & SSIM & LPIPS & Params (M) \\
        \hline
        Freqformer w/ crop & 23.13 & 0.9220 & 0.0695 & 2.695 \\
        Freqformer w/ crop \& tile & 25.90 & 0.9363 & 0.0486 & 2.695 \\
        Freqformer w/ resize & 29.90 & 0.9568 & 0.0291 & 2.695 \\
        \hline
        ESDNet-L w/ crop & 23.01 & 0.9192 & 0.0736 & 10.623 \\
        ESDNet-L w/ crop \& tile & 25.35 & 0.9295 & 0.0575 & 10.623 \\
        ESDNet-L w/ resize & 30.39 & 0.9570 & 0.0277 & 10.623 \\
        \bottomrule
    \end{tabular}}
    \caption{Low-branch resizing.}
    \vspace{-4mm}
    \label{tab:resize}
\end{table}

\textbf{Low-frequency branch training via resize.} As we proposed in our method, the resize operation contains many benefits during the low-branch training, which are demonstrated in Tab.~\ref{tab:resize}. We also train an ESDNet-L (10M) to verify our findings in the same setting. Outcomes show that the crop-training strategy prevents the model from leveraging the low-frequency component’s inherent robustness to resizing and hinders effective global information aggregation for color correction, resulting in significant degraded performance. We further introduce a crop-tile inference strategy (crop the low-frequency part to 512$\times$512 patches with 32 overlap and then assemble them) for the crop-training model, but it is still inferior to the resize-strategy.

\textbf{Learnable Frequency Composition Transform and dual-branch training.} In Tab.~\ref{tab:e2e}, we demonstrate the importance of our frequency decomposition and two-branch training strategy. As shown, ESDNet and Freqformer, using direct modeling without frequency decomposition (w/o F.D.), perform worse than the two-branch approach. Moreover, incorporating a learnable FCT improves fusion consistency and pattern coherence with few parameters.

\begin{table}[htbp]
    \centering
    \vspace{-2mm}
    \setlength{\tabcolsep}{3pt}
    \resizebox{0.7 \textwidth}{!}{
    \begin{tabular}{l|ccc|c}
        \toprule
        Model & PSNR & SSIM & LPIPS & Params (M) \\
        \hline
        ESDNet w/ learnable FCT & 25.20 & 0.8511 & 0.1244 & 12.567 \\
        ESDNet w/ fixed FCT & 25.11 & 0.8469 & 0.1296 & 11.868 \\
        ESDNet-L (w/o F.D.) & 24.88 & 0.8440 & 0.1301 & 10.623 \\
        \hline
        Freqformer w/ learnable FCT & 25.26 & 0.8518 & 0.1253 & 6.065 \\
        Freqformer w/ fixed FCT & 25.13 & 0.8478 & 0.1298 & 5.876 \\
        Freqformer-L (w/o F.D.) & 23.92 & 0.8409 & 0.1284 & 5.851 \\
        \bottomrule
    \end{tabular}}
    \caption{Effects of learnable FCT and frequency decomposition.}
    \label{tab:e2e}
    \vspace{-5mm}
\end{table}

\textbf{Freqformer component ablation.} As shown in Tab.~\ref{tab:compo}, hierarchical fusion (denoted as H.F. in the Table) improves performance with minimal computational and parameter overhead. Replacing RDDB with deeper transformer layers shows that RDDB is more efficient and outperforms the pure transformer design with faster speed. We further replace SCP (1$\times$1 convolution and a 3$\times$3 depth-wise convolution) with a full 3$\times$3 convolution that uses twice as many parameters, yet we observe no performance improvement and even a speed reduction.
We also explored the sophisticated DAT~\cite{chen2023dual} block as a baseline, but it underperforms our SA-CA, which more effectively aggregates local features. 
\begin{table}[htbp]
    \centering
    \vspace{-2mm}
    \resizebox{0.7 \textwidth}{!}{
    \begin{tabular}{l|cc|cc}
        \toprule
        Model (High branch) & PSNR & SSIM & Params (M) & Speed (s) \\
        \hline
        Freqformer & 29.20 & 0.8805 & 3.181 & 0.128 \\
        Freqformer (w/o H.F.) & 29.15 & 0.8804 & 2.997 & 0.126 \\
        Freqformer (Deeper, w/o RDDB) & 29.19 & 0.8801 & 2.944 & 0.140 \\
        Freqformer (Full Conv. SCP) & 29.18 & 0.8804 & 6.451 & 0.187 \\
        Freqformer (DAT Arch.) & 29.10 & 0.8793 & 3.107 & 0.196 \\
        \bottomrule
    \end{tabular}}
    \caption{The ablation of the Freqformer's architecture.}
    \vspace{-6mm}
    \label{tab:compo}
\end{table}

\vspace{-2mm}
\section{Conclusion}
\vspace{-2mm}
In this paper, we propose Freqformer, a novel demoiréing framework leveraging frequency separation to address complex moiré patterns. By introducing a tailored frequency decomposition strategy and a learnable FCT module, we effectively disentangle moiré artifacts into high-frequency textures and low-frequency color distortions. Furthermore, we design a dual-branch architecture with asymmetric learning strategies: crop-based training for high frequencies and resize-based training for low frequencies. Additionally, we introduce a lightweight yet effective SA-CA module to boost moiré-sensitive feature learning while maintaining efficiency, making our model suitable for deployment on high-resolution images and edge devices. This study highlights how integrating frequency decomposition, learnable fusion, and spatial-channel attention advances image demoiréing, offering promising directions for future exploration.

\section*{Acknowledgements}
This work is supported by the National Natural Science Foundation of China (62501386, 625B2116), CCF-Tencent Rhino-Bird Open Research Fund, and CAAI-Tencent Rhino-Bird Open Research Fund. This work is also sponsored by AI Hundred Schools Program and is carried out using the Ascend AI technology stack.

%
%
\bibliographystyle{splncs04}
\bibliography{main}
\end{document}